\documentclass[10pt,twocolumn,letterpaper]{article}

\usepackage{cvpr} 
 
\usepackage{graphicx}
\usepackage{amsmath}
\usepackage{amssymb}
\usepackage{booktabs}
\usepackage{xcolor}
\usepackage{colortbl}
\usepackage{adjustbox}
\usepackage{dsfont}
\usepackage{multirow}

\usepackage{algorithm}% http://ctan.org/pkg/algorithms
\usepackage{algpseudocode}% http://ctan.org/pkg/algorithmicx

\usepackage[pagebackref,breaklinks,colorlinks]{hyperref}

\begin{document}

\title{Weakly supervised segmentation with cross-modality equivariant constraints}

\author{Gaurav Patel$^1$, Jose Dolz$^{2}$\\
\\ \vspace{1mm}
\textit{$^1$ Purdue University, West Lafayette, IN 47907, USA}\\
\textit{$^2$ \'{E}cole de Technologie Sup\'{e}rieure, Montreal, QC H3C 1K3, Canada}\\
{\tt\small \{pate1332, gpatel10\}@purdue.edu., jose.dolz@etsmtl.ca}
}


\maketitle

\begin{abstract}
Weakly supervised learning has emerged as an appealing alternative to alleviate the need for large labeled datasets in semantic segmentation. Most current approaches exploit class activation maps (CAMs), which can be generated from image-level annotations. Nevertheless, resulting maps have been demonstrated to be highly discriminant, failing to serve as optimal proxy pixel-level labels. We present a novel learning strategy that leverages self-supervision in a multi-modal image scenario to significantly enhance original CAMs. In particular, the proposed method is based on two observations.  First, the learning of fully-supervised segmentation networks implicitly imposes equivariance by means of data augmentation, whereas this implicit constraint disappears on CAMs generated with image tags. And second, the commonalities between image modalities can be employed as an efficient self-supervisory signal, correcting the inconsistency shown by CAMs obtained across multiple modalities. To effectively train our model, we integrate a novel loss function that includes a within-modality and a cross-modality equivariant term to explicitly impose these constraints during training. In addition, we add a KL-divergence on the class prediction distributions to facilitate the information exchange between modalities which, combined with the equivariant regularizers further improves the performance of our model. Exhaustive experiments on the popular multi-modal BraTS and prostate DECATHLON segmentation challenge datasets demonstrate that our approach outperforms relevant recent literature under the same learning conditions.
\end{abstract}

\section{Introduction}

Semantic segmentation is of pivotal importance in medical image analysis, as it serves as precursor for many downstream tasks, such as diagnosis, treatment or follow-up. With the advent of deep learning, state-of-the-art in medical image segmentation is dominated by these models \cite{chen2018voxresnet,dolz20183d,milletari2016v}, which largely outperform more traditional methods. %and can be applied to multiple problems. 
Nevertheless, these models require large labeled datasets, which is a cumbersome process and require user-expertise. Thus, novel learning approaches that can alleviate the need of large labeled datasets are highly desirable.

Weakly supervised segmentation (WSS) has appeared as an appealing alternative to fully-supervised learning to overcome the scarcity on annotations. In this scenario, supervision is typically given in the form of image tags\cite{pinheiro2015image}, points \cite{bearman2016s}, scribbles \cite{lin2016scribblesup}, bounding boxes \cite{kervadec2020bounding,rajchl2016deepcut} or global target information, such as object size  \cite{kervadec2019constrained,kervadec2019curriculum,pathak2015constrained,peng2020discretely}. A common strategy is to use image-level labels to derive pixel-wise class activation maps (CAMs) \cite{zhou2016learning}, which serve to identify object regions in the image. The resulting maps, however, are highly discriminative on the target object, while fail to correctly locate background regions, resulting in either under- or over-segmentations. To address this issue, different alternatives to improve the initial CAMs have been proposed. While combining CAMs with saliency maps is a popular choice \cite{fan2020learning,oh2017exploiting}, other approaches resort to iterative region mining strategies \cite{wei2017object}.

\begin{figure}[t!]
\begin{centering}
\includegraphics[width=1\linewidth]{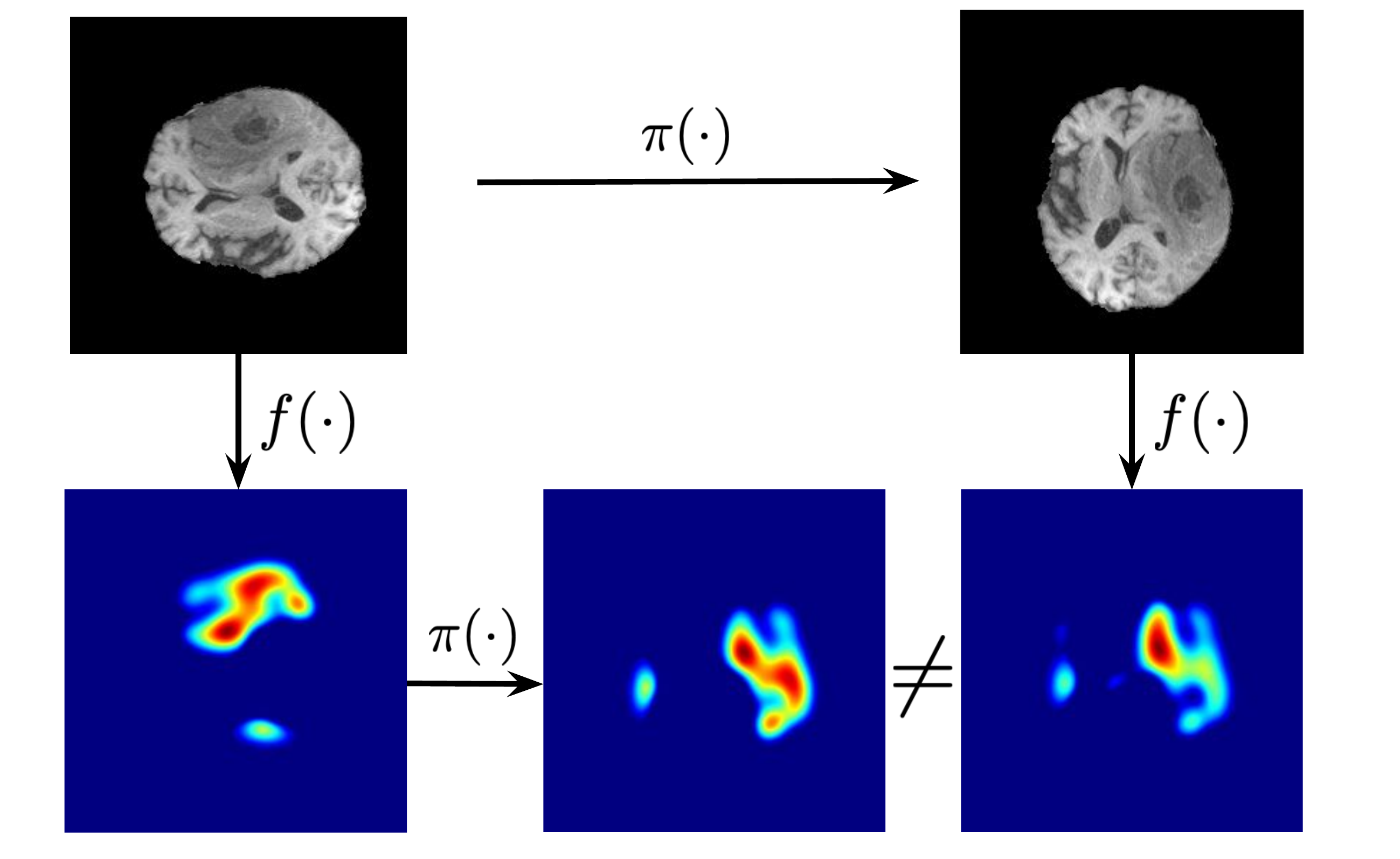}
\par\end{centering}
\caption{Visual example demonstrating non-equivariance to spatial transformations in regular CNNs used to generate class-activation maps, i.e.,  $f(\pi(\cdot))\neq\pi(f(\cdot)).$ In this example, $f(\cdot)$ denotes the function to generate the class-activation maps, whereas $\pi(\cdot)$ represents the set of potential spatial affine transformations. In pixel-wise recognition tasks, such as segmentation, a CNN equivariant to spatial transformations is expected to follow $f(\pi(\cdot))=\pi(f(\cdot))$. In other words, spatially modifying the input is expected to result in the same modification in the output, even though this is not observed in standard CNNs.} \label{fig:Class-Activation-maps2}
\end{figure}

\begin{figure}[t!]
\begin{centering}
\includegraphics[width=1\linewidth]{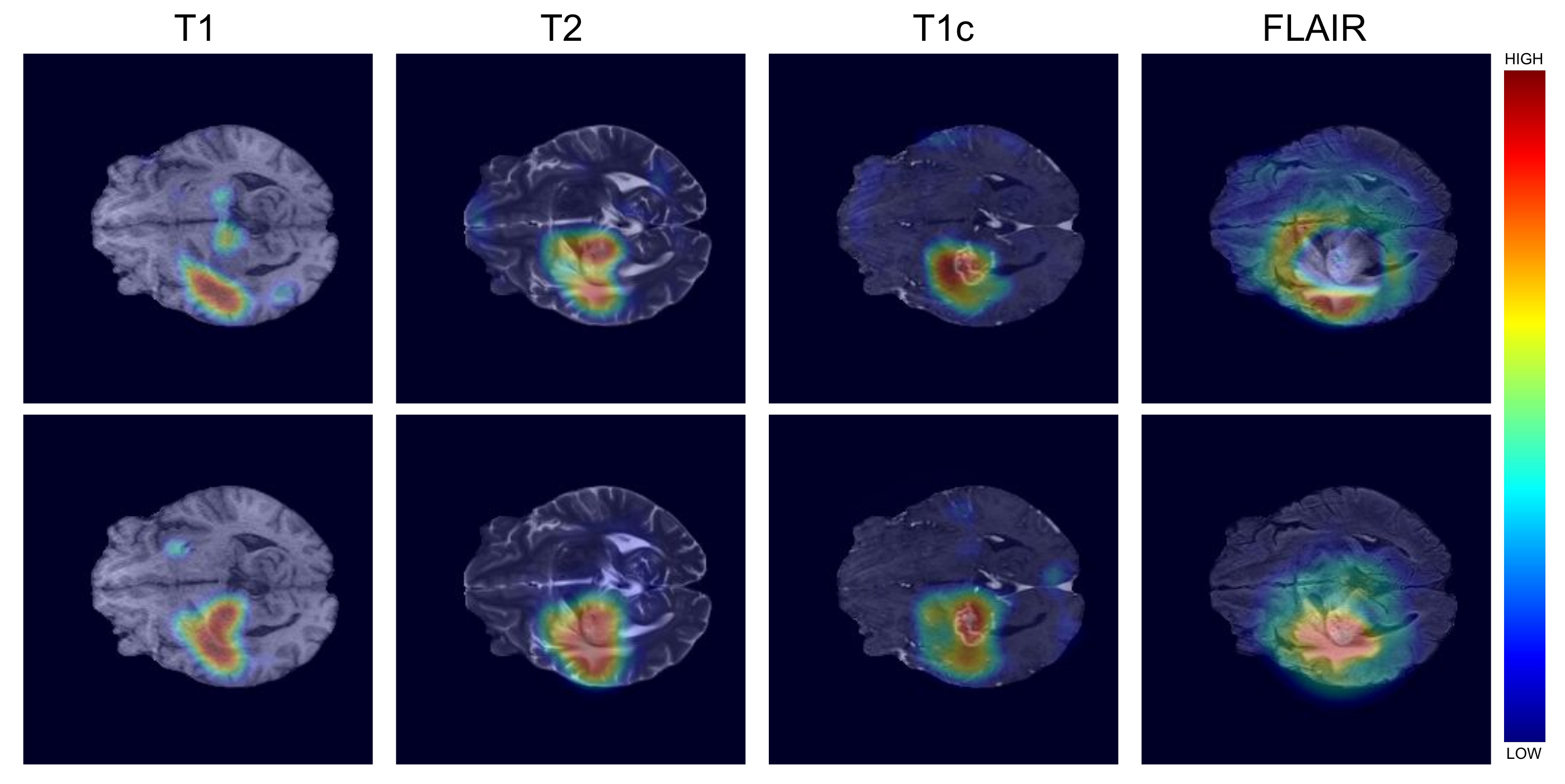}
\par\end{centering}
\caption{Class activation maps (CAMs) obtained for different image modalities (represented as columns). The baseline CAM \cite{zhou2016learning} is depicted in the \textit{top} row, whereas CAMs obtained by the proposed model are shown in the \textit{bottom} row. We can observe that CAMs generated by our cross-modality approach are more consistent across modalities. \label{fig:Class-Activation-maps}}
\end{figure}

Despite the wide adoption of CAMs for weakly supervised segmentation, and to the best of our knowledge, two important facts have been overlooked in the current literature.  First,  data augmentation is widely employed in the training of fully-supervised segmentation models.  During this stage, several affine transformations are applied equally to both the input images and their corresponding pixel masks, which implicitly introduces an equivariant constraint to the model. While this is not an issue in fully-supervised segmentation models, this implicit constraint is lost in the resulting CAMs, as they are obtained from the image level labels. In particular, the global average pooling (GAP) operation integrated to generate the CAMs makes the spatial information to be lost, resulting in different CAMs across transformations. %This results in large inconsistencies on CAMs generated on the same image with different affine transformations.
This is evidenced in large inconsistencies on CAMs found across different affine transformations of a given image, which suggests that regular CNNs used to generate CAMs are not inherently equivariant (See Fig. \ref{fig:Class-Activation-maps2} for a visual explanation of this phenomenon). Nevertheless, only the recent work in \cite{wang2020self} exploited this observation to enhance the generated CAMs. And second, as shown in Figure \ref{fig:Class-Activation-maps}, CAMs generated from the same image across different modalities are also inconsistent, despite they represent the same region. As highlighted in these two figures, disparity between the CAMs is further magnified across multi-modal images.

Inspired by these limitations, we present in this paper a novel learning strategy for weakly supervised segmentation in a multi-modal scenario. Particularly, to leverage the complimentary information across image modalities we integrate three different terms in the learning objective. First, explicit intra-modality constraints enhance individual CAMs by forcing them to be similar across spatial image transformations. Second, in order to facilitate the information exchange across multiple networks, we integrate a Kullback-Leibler (KL) term on the networks outputs. Nevertheless, as we show in our experiments, this cross-modality information flow does not necessarily guarantee improvements on the object localization. To overcome this issue, we introduce a cross-modality equivariant constraint, which applies consistency regularization on the CAMs generated %from transformed images
across modalities. This regularization provides a mechanism of self-supervision which leads to enhanced CAMs, as we show in Figure \ref{fig:Class-Activation-maps}. Thus, the main contributions of this work can be summarized as:

\begin{itemize}
    \item We propose a novel and effective weakly supervised segmentation strategy in the multi-modal imaging scenario.
    \item The proposed learning strategy leverages multi-modal images to generate enhanced CAMs under image-level supervision. In particular, we introduce intra-modality and cross-modality equivariant constraints, %in the multi-modal scenario, 
which guide the multi-modal learning, leading to better object localizations. 

    \item We conduct extensive experiments on the popular BRaTS dataset, demonstrating that the proposed consistency regularization terms bring a substantial gain on performance over the current state-of-the-art on weakly supervised segmentation. In addition, the generalization capabilities of our approach are demonstrated in the multi-modal prostate dataset from the DECATHLON segmentation challenge.
    \item Furthermore, we also provide insights on the source of the improvement from the proposed method.
\end{itemize}

\section{Related Work}

\subsection{Weakly supervised segmentation}
In contrast to fully supervised learning, deep networks trained under the weakly supervised learning paradigm make use of weak labels for training guidance. We can broadly categorize these methods as data-driven \cite{dai2015boxsup,khoreva2017simple,kolesnikov2016seed,lin2016scribblesup,pinheiro2015image} and knowledge-drive approaches \cite{li2017instance,pathak2015constrained,remez2018learning,hou2017deeply,zhang2017curriculum}. In the first category supervision can come in the form of image-level labels \cite{kolesnikov2016seed,pinheiro2015image}, scribbles \cite{lin2016scribblesup} or bounding boxes \cite{dai2015boxsup,khoreva2017simple}, for example. On the other hand, the supervision in the second group is given as a prior knowledge, such as target size \cite{pathak2015constrained,zhang2017curriculum}, location \cite{remez2018learning} or existence of high contrast between background and foreground, i.e., saliency \cite{hou2017deeply,li2017instance}. Note that both data-driven and knowledge-driven information can be used jointly in a single approach \cite{oh2017exploiting,tang2018regularized}. 

A popular strategy, however, is to resort to image-level labels to generate  class activation map (CAM) \cite{selvaraju2017grad}, which are later employed as pixel-wise pseudo-masks to train segmentation networks, mimicking full supervision. As original CAMs are highly discriminative, resulting in under segmentations of the object of interest, a large body of literature has focused on enhancing these initial CAMs. A common choice to expand the initial seeds is to integrate saliency information   \cite{fan2020learning,kolesnikov2016seed,oh2017exploiting}. For example, \cite{kolesnikov2016seed} exploited a seed-expand-constraint framework where class-independent saliency maps\cite{simonyan2013deep} were used to refine initial CAMs. Other approaches rely on iterative mining strategies \cite{wei2017object,wang2018weakly}, which progressively expand object regions to eventually constitute a more complete object usable for semantic segmentation. More recently, and closely related to our work, \cite{wang2020self} proposed an equivariant attention mechanism which serves as a regularizer for the CAMs in color images.   

Despite the wide use of these methods in computer vision, the literature on medical image segmentation remains scarce. DeepCut \cite{rajchl2016deepcut}, which is strongly inspired by \cite{papandreou2015weakly}, resorts to the popular GrabCut \cite{rother2004grabcut} approach to generate image-proposals from bounding box annotations. These proposals are later used as pseudo-masks to train a segmentation network. More recently, image-level labels have also been leveraged to generate initial seeds, i.e., CAMs \cite{nguyen2019novel,ouyang2019weakly,wu2019weakly}, % I removed schumacher2020weakly to reduce space
which, similar to \cite{rajchl2016deepcut}, serve as pseudo-masks in a later step. Knowledge-driven approaches prevail in the medical domain, where the prior-knowledge is typically integrated as an augmented loss function \cite{Jia2017,kervadec2019constrained,kervadec2019curriculum,kervadec2020bounding}. For example, \cite{kervadec2019constrained} force the segmentation output to satisfy a given region size. This term is integrated as a differentiable penalty that enforces inequality constraints directly in the learning objective. 

Nevertheless, these approaches have overlooked the complementary information contained in multi-modal data which, as we will demonstrate, substantially helps to obtain more meaningful regions.

\subsection{Self-training}

Self-training has recently attracted considerable attention as a proxy to learn robust representations. In this learning paradigm, labels can be obtained from unlabeled images via pre-text tasks. Classical pre-text tasks include predicting image orientation \cite{gidaris2018unsupervised} or relative position prediction \cite{doersch2015unsupervised}, solving jigsaw puzzles \cite{noroozi2016unsupervised}, image inpainting \cite{pathak2016context}, colorization \cite{larsson2016learning} and many others \cite{chaitanya2020contrastive,doersch2017multi,dosovitskiy2014discriminative,zhang2017split}. 
More recently, equivariance has been employed to impose semantic consistency, either at keypoints \cite{thewlis2017unsupervised}, class activations \cite{wang2019self}, feature representations  \cite{sahasrabudhe2020self} or the network outputs \cite{hung2019scops}. 

Nevertheless, a main limitation of these works is that equivariance is enforced across affine transformations of the same image \cite{li2020transformation,sahasrabudhe2020self,wang2020self} or between virtually generated versions \cite{hung2019scops}. For instance, \cite{hung2019scops} further apply an appearance perturbation (e.g., color jittering) to the input image before the affine transformation. The limitation of this approach is that, even though the images present appearance differences, these are not complimentary as in the case of multiple MRI modalities. Contrary to these works, we advocate that enforcing equivariant constraints across multiple image modalities results in representations that are more robust to image variations. This is particularly important in medical %multi-modal 
imaging, where each modality highlights different tissue properties.

\subsection{Multi-modal segmentation}

Combining multiple image modalities has been extensively employed to learn more powerful representations in multi-modal fully-supervised scenarios. In this context, early and late fusion are commonly used.  In particular, early fusion involves concatenating the multiple modalities in the input of the network, each representing an individual channel  \cite{moeskops2016automatic,zhang2015deep}. On the other hand, late fusion promotes processing modalities individually, whose features are fused in a later stage \cite{li20183d,nie2016fully}, typically at the last convolutional layers. In addition, there exist recent works that present more sophisticated multi-modal fusion strategies. These models include leveraging dense connectivity \cite{dolz2018hyperdense} or multi-scale fusion strategies \cite{li2019mman}, among others. However, despite the increasing interest seen in multi-modal segmentation, the literature has mostly focused on fully-supervised tasks. Thus, our work contrasts to prior literature, as we aim at leveraging multi-modal information under the weakly supervised learning paradigm.

\section{Methodology}

\begin{figure*}[t!]
\includegraphics[width=1\linewidth]{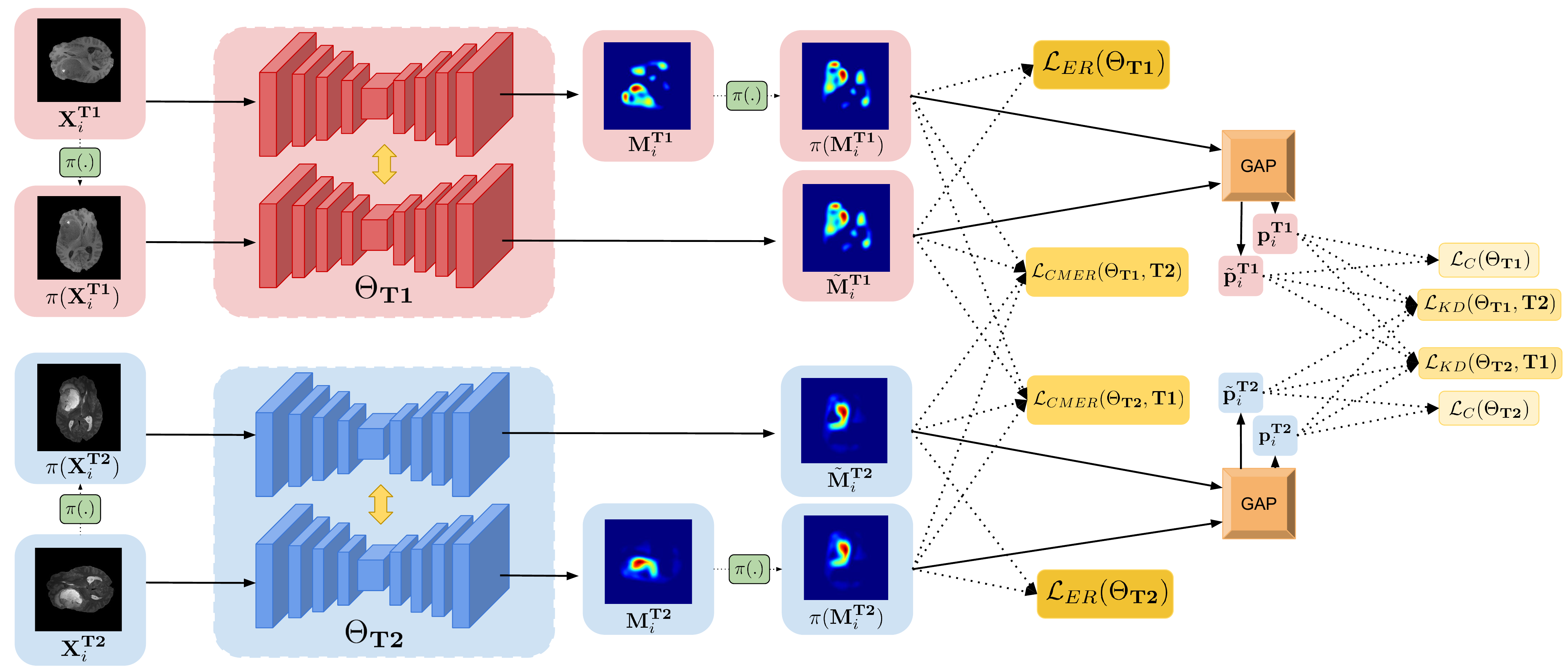}

\caption{Overview of the proposed weakly supervised learning strategy for multi-modal images under equivariant constraints. Particularly, we show the pipeline for the two-modalities setting, i.e., T1 and T2. The Each of the blocks processing single modalities are depicted in colors (red and blue). Then, we employ different shades of yellow to represent the multiple loss terms employed for training. Our model enhances the class activation maps in multi-modal imaging by coupling an intra-modality and a cross-modality equivariant constraint (Section \ref{ssec:self}). Furthermore, the KL terms  ($\mathcal{L}(\Theta_{m_{1}},m_{2})$ and $\mathcal{L}(\Theta_{m_{2}},m_{1})$) facilitate the information exchange between modalities. Note that the GAP module refers to a standard global average pooling (GAP) layer at the end of each model.} 
\label{Pipeline}
\end{figure*}

\subsection{Formulation}

Let us denote a set of $N$ training images as $\mathcal{D}=\{(\{\mathbf{X}_{i}^{m_k}\}_{k=1}^{K},\mathbf{y}_{i})\}_{i=1}^{N}$
where $\mathbf{X}_{i}^{{m_k}}\in\mathbb{R^{\mathit{\Omega_{i}^{m_k}}}}$ is
an image belonging to modality $m_k$, $K$ denotes the total number of modalities, $\mathit{\Omega_{i}^{{m_k}}}$ denotes the spatial image dimension and $\mathbf{y}_{i}\in\{0,1\}^{C}$ its corresponding one-hot encoded class label, with $C$ indicating the number of distinct classes. Furthermore, we denote $\mathcal{P}=\{\{\mathbf{p}_{i}^{m_{k}}\}_{k=1}^{K}\}_{i=1}^{N}$ as the set of corresponding vectors of softmax probabilities, obtained from the final classification layer, where $\mathbf{p}_{i}^{m_{k}} = f_{\Theta_{m_{k}}}(\mathbf{X}_{i}^{{m_k}})\in[0,1]^{C}$, being $f_{\Theta_{m_{k}}}(\cdot)$ a convolutional neural network parameterized by $\Theta_{m_{k}}$. Additionally, following the literature in weakly supervised segmentation, we can obtain the corresponding image class activation maps (CAMs\footnote{Note that in our implementation we use the single-step alternative to obtaining the CAM as in \cite{zhang2018adversarial}, which avoids the additional weight multiplication of the fully-connect layers with the feature maps as in the original CAMs \cite{zhou2016learning}.}) from samples in $\mathcal{D}$, resulting in the set $\mathcal{M}=\{\{\mathbf{M}_{i}^{{m_k}}\}_{k=1}^{K}\}_{i=1}^{N}$. In this set, $\mathbf{M}_{i}^{{m_k}}\in[0,1]^{\mathit{\Omega}_{i}^{{m_k}}\times C-1}$\footnote{The class activation map corresponding to the background class is not included, therefore this results in a vector of dimension $C-1$.} represents the max-normalized CAM of the $i^{th}$ sample and modality ${m_k}$.

\subsection{Self-supervision with Transformation Equivariance}
\label{ssec:self}
 
Many WSS %weakly supervised segmentation 
methods jointly leverage classification and segmentation models by training a classification network first and then keeping part of the network as features extractor to tackle the segmentation task. Take for example a segmentation neural network $f_{\Theta_s}(\cdot)$, which produces a pixel-mask segmentation $\hat{\mathbf{Y}}=f_{\Theta_s}(\mathbf{X})$ for a given image $\mathbf{X}$. According to the literature in WSS, this model can be transformed into a classification network by simply adding a pooling layer, $\hat{\mathbf{y}}=Pool(f_{\Theta_s}(\mathbf{X}))$, as they are built on the assumption that learned parameters are equivalent. Nevertheless, despite the commonalities between these networks, there exist underlying differences that can be exploited in the context of semantic segmentation. Particularly, while the former is transformation invariant --random image transformations should lead to the same prediction--, the latter is transformation equivariant. This means that for a random spatial image transformation $\pi(\cdot)$, the pixel-wise prediction should be affected identically, $f_{\Theta_s}(\pi(\mathbf{X}))=\pi(f_{\Theta_s}(\mathbf{X}))$. This motivates the integration of equivariance properties in the learning objective, which can result in strong regularizers that allow an explicit mechanism to include equivariant constraints. 

We first generate an augmented training set where each image in $\mathcal{D}$ follows a series of spatial transformations $\pi(\cdot)$, resulting in $\mathcal{D}_{\pi}=\{\{\pi(\mathbf{X}_{i}^{m_{k}})\}_{k=1}^{K},\mathbf{y}_{i})\}_{i=1}^{N}$. Note that each image follows the same transformation across the $K$ modalities. Then, we generate their corresponding CAMs, which are denoted as $\mathcal{M}_{\pi}=\{\{\tilde{\mathbf{M}}_{i}^{m_{k}}\}_{k=1}^{K}\}_{i=1}^{N}$, and obtain the softmax vector probabilities on the transformed images, $\mathcal{P}_{\pi}=\{\{\tilde{\mathbf{p}}_{i}^{m_{1}}\}_{k=1}^{K}\}_{i=1}^{N}$. For simplicity and the ease of explanation we formulate
our method in a bi-modal setup, with image modalities $m_{1}$ and
$m_{2}$ operated on networks with weights $\Theta_{m_{1}}$ and $\Theta_{m_{2}}$,
respectively, and later generalize to $K$ modalities.

\paragraph{\textbf{Within modalities equivariance}}Equivariant regularizers on single modalities have been recently explored in weakly-supervised \cite{wang2020self} and self-supervised \cite{sahasrabudhe2020self} learning. %In these works, the explicit equivariant constraint is formally defined as:
This explicit constraint can be formally defined as:
\begin{equation}
\mathcal{R_{\mathit{E}}}=\parallel\pi(f(\mathbf{X}))-f(\pi(\mathbf{X}))\parallel_{2}^{2}\label{eq:1}
\end{equation}

where $f(\cdot)$ could be the same network \cite{sahasrabudhe2020self} or an expanded version consisting in a shared-weight Siamese structure with two branches \cite{wang2020self}. Following these works, we define the first of our objective terms, a transformation
equivariance regularization loss, which enforces transformation consistency within the same modality:

\begin{equation}
\label{eq:ER}
\mathcal{L}_{ER}(\Theta_{m_{1}})=\parallel\pi(\mathbf{M}_{i}^{m_{1}})-\tilde{\mathbf{M}}_{i}^{m_{1}}\parallel_{2}^{2}
\end{equation}

% Cross Modal Distillation for Supervision Transfer
\paragraph{\textbf{Information exchange across multiple networks}}In addition to single-modality equivariance, we aim at facilitating information flow across networks, each corresponding to individual modalities $m_1$ and $m_2$. To achieve this, we first enforce the softmax likelihood from the first network, $\mathbf{p}_{i}^{m_{1}}$, to match the posterior probability from the second network, $\mathbf{p}_{i}^{m_{2}}$, for both original and transformed images. This is similar to \cite{gupta2016cross,zhang2018deep}, where cross-modal distillation was employed as a supervisory signal in the context of object region detection and classification. The resulting objective can be formulated as a Kullback-Leibler (KL) divergence term:

\begin{equation}
\mathcal{L}_{KD}(\Theta_{m_{1}},m_{2})=\frac{1}{2}(\mathcal{D}_{KL}(\mathbf{p}_{i}^{m_{2}}||\mathbf{p}_{i}^{m_{1}})+\mathcal{D}_{KL}(\tilde{\mathbf{p}}_{i}^{m_{2}}||\tilde{\mathbf{p}}_{i}^{m_{1}}))
\label{eq:kl}
\end{equation}

where $\mathcal{D}_{KL}(\mathbf{p}||\mathbf{q})=\mathbf{p}^{\top}\log(\frac{\mathbf{p}}{\mathbf{q}})$. We resort to $\mathcal{L}_{KD}(\cdot,m_{k})$ to emphasize that the knowledge is distilled from the \textit{k}-th modality. Note that our KL divergence loss is asymmetric and thus different for each network, which explains the need of using two asymmetric terms. Instead, we could have used a symmetric Jensen-Shannon Divergence loss, but \cite{zhang2018deep} observed that empirically employing either a symmetric or an asymmetric KL loss does not make any difference.

Nevertheless, information exchange on the predicted likelihood
distribution does not guarantee improvement on object localization, as we will show in the experimental section. Thus, inspired from \cite{Zagoruyko2017AT} we seek to mine complementary %activation
information present in CAMs derived from multiple modalities. To achieve this, we impose equivariance constraints on multi-modal CAMs, which we refer to as the Cross-Modal Equivariance Regularization (CMER) loss, defined as: 

\begin{multline}
\mathcal{L}_{CMER}(\Theta_{m_{1}},m_{2})=\frac{1}{2}(\parallel\pi\left(\frac{\mathbf{M}_{i}^{m_{2}}}{\parallel\mathbf{M}_{i}^{m_{2}}\parallel_{2}}\right)-\frac{\tilde{\mathbf{M}}_{i}^{m_{1}}}{\parallel\tilde{\mathbf{M}}_{i}^{m_{1}}\parallel_{2}}\parallel_{2}^{2}\\
+\parallel\frac{\tilde{\mathbf{M}}_{i}^{m_{2}}}{\parallel\tilde{\mathbf{M}}_{i}^{m_{2}}\parallel_{2}}-\pi \left(\frac{\mathbf{M}_{i}^{m_{1}}}{\parallel\mathbf{M}_{i}^{m_{1}}\parallel_{2}}\right)\parallel_{2}^{2})\label{eq:cmer}
\end{multline}

\subsection{Classification loss}

In the context of this work, we only have access to image-level labels as supervisory signals. To leverage this information, we integrate a global average pooling (GAP) layer at the end of the model (Fig. \ref{Pipeline}). The role of this layer is to aggregate the CAMs derived from the original and transformed images at each branch into the prediction vectors $\mathbf{p}^{m_k}$ and $\mathbf{\tilde{p}}^{m_k}$, respectively, for image classification. To train the network we resort to the standard cross-entropy loss $\mathcal{L}_{CE}(\mathbf{p},\mathbf{y})=-\mathbf{y}^{\top}\log(\mathbf{p})$, where $\mathbf{p}$ and $\mathbf{y}$ are the column vectors of length
$C$ of the predicted softmax probabilities and the one-hot encoded ground-truth,
respectively. As we employ a Siamese structure to impose the self-equivariance, we extend the image-level supervision to both the branches. Thus, for the network processing the \textit{k}-th modality, the classification loss becomes: 

\begin{multline}
\mathcal{L}_{C}(\Theta_{m_{k}})=\frac{1}{2}(\mathcal{L}_{CE}(\mathbf{p}_{i}^{m_{k}},\mathbf{y}_{i})+\mathcal{L}_{CE}(\tilde{\mathbf{p}}_{i}^{m_{k}},\mathbf{y}_{i})).
\label{eq:class_loss}
\end{multline}

\subsection{Global objective}
Finally, the overall loss function for our model can be defined as $\mathcal{L} = \mathcal{L}(\Theta_{m_{1}}) + \mathcal{L}(\Theta_{m_{2}})$, where each terms represents the loss for each network. Particularly, for the model parameterized by $\Theta_{m_{1}}$, this loss can be defined as:

\begin{multline}
\mathcal{L}(\Theta_{m_{1}})=\mathcal{L}_{C}(\Theta_{m_{1}})+\lambda_{KD}\mathcal{L}_{KD}(\Theta_{m_{1}},m_{2})\\
+\lambda_{E}(t)(\mathcal{L}_{ER}(\Theta_{m_{1}})+\mathcal{L}_{CMER}(\Theta_{m_{1}},m_{2}))
\end{multline}

where $\lambda_{KD}$ and $\lambda_{E}(t)$ balance the importance of each term in the learning objective. Similarly, for the network parameterized with $\Theta_{m_{2}}$, the
overall loss function is:
\begin{multline}
\mathcal{L}(\Theta_{m_{2}})=\mathcal{L}_{C}(\Theta_{m_{2}})+\lambda_{KD}\mathcal{L}_{KD}(\Theta_{m_{2}},m_{1})\\
+\lambda_{E}(t)(\mathcal{L}_{ER}(\Theta_{m_{2}})+\mathcal{L}_{CMER}(\Theta_{m_{2}},m_{1}))
\end{multline}

\subsection{Generalization to K-modalities}
\label{ssec:k-mod}

We now generalize our learning objective to $K$ modalities, which involves $K$ networks learning simultaneously. For a particular network $\Theta_{m_{k}}$ processing the modality $m_k$, the loss function can be defined as:
 
\begin{multline}
\mathcal{L}(\Theta_{m_{k}})=\mathcal{L}_{C}(\Theta_{m_{k}})\\
+\lambda_{KD}(\frac{1}{K-1}\sum_{l=1,l\neq k}^{l=K}\mathcal{L}_{KD}(\Theta_{m_{k}},m_{l}))\\
+\lambda_{E}(t)(\mathcal{L}_{ER}(\Theta_{m_{k}})+\frac{1}{K-1}\sum_{l=1,l\neq k}^{l=K}\mathcal{L}_{CMER}(\Theta_{m_{k}},m_{l}))\label{eq:100}
\end{multline}

Our learning strategy is detailed in Algo. \ref{alg:main_algorithm}.

\begin{algorithm}
%\SetAlgoLined
\caption{Training algorithm}
\label{alg:main_algorithm} 
\begin{algorithmic}[1] 
\Require{Training dataset $\mathcal{D}$}
\State {$K$ = Number of image modalities}
\State {$\Pi$ = Set of transformations}
\State {$T$ = Total number of Epochs}

\For {$k$ in $[1,K]$}
	\State {Initialize $\Theta_{m_{k}}$} 
\EndFor   

\For{$t$ in $[1,T]$ }
	\For{every minibatch $B$ in $\mathcal{D}$}
		\State {$\pi\gets\pi\sim\Pi$}
		\State {$\mathcal{M}\gets\{\{\mathbf{M}_{i}^{m_{k}}\}_{k=1}^{K}\}_{i\in B}$}
		\State {$\mathcal{P}\gets\{\{\mathbf{p}_{i}^{m_{k}}\}_{k=1}^{K}\}_{i\in B}$}
		\State {$\mathcal{D}_{\pi}\gets\{\{\pi(\mathbf{X}_{i}^{m_{k}})\}_{k=1}^{K},\mathbf{y}_{i})\}_{i\in B}$}
		\State {$\mathcal{M}_{\pi}\gets\{\{\tilde{\mathbf{M}}_{i}^{m_{k}}\}_{k=1}^{K}\}_{i\in B}$}
		\State {$\mathcal{P}_{\pi}\gets\{\{\tilde{\mathbf{p}}_{i}^{m_{k}}\}_{k=1}^{K}\}_{i\in B}$}
		\For {$k$ in $[1,K]$}
		    \State Compute $\mathcal{L}_{C}(\Theta_{m_{k}})$, $\mathcal{L}_{ER}(\Theta_{m_{k}})$,
		    \State $\mathcal{L}_{KD}(\Theta_{m_{k}},m_{l\neq k})$,
		    $\mathcal{L}_{CMER}(\Theta_{m_{k}},m_{l\neq k})$.
		    \State Compute $\mathcal{L}(\Theta_{m_{k}})$
			\State {$loss\gets\frac{1}{\mid B\mid}\sum_{i\in B}\mathcal{L}(\Theta_{m_{k}})$}
			\State {Compute gradients of $loss$ w.r.t $\Theta_{m_{k}}$}
			\State {Update $\Theta_{m_{k}}$ using the optimizer}
		\EndFor
	\EndFor
\EndFor
\end{algorithmic}
\end{algorithm}

\section{Experimental setting}
\subsection{Dataset}
We benchmark the proposed method in the context of multi-modal brain tumor segmentation in MR images and prostate segmentation in MR-T2 and apparent diffusion coefficient (ADC) maps.

\paragraph{\textcolor{black}{\textbf{Brain tumor segmentation}}}\textcolor{black}{For this task}, we use the popular BraTS 2019 dataset \cite{Brats_bakas2017advancing,bakas2018identifying,Brats_menze2014multimodal}. This dataset contains 335 multi-modal scans with their corresponding expert segmentation masks. More concretely, the scans are composed of four modalities, which include T1, T1c, T2 and FLAIR. The images were re-sampled to an isotropic 1.0 $mm$ voxel spacing, skull-striped and co-registered by the challenge organizers. 
In the context of this work, we consider a binary segmentation class, i.e., healthy vs non-healthy targets. %, to avoid the inter-class inference on the generated CAMs. 
Thus, we merge the different tumour classes into a common non-healthy class. \textcolor{black}{We would like to emphasize that the proposed approach is not limited to binary scenarios, but to the nature of the available data. Class activation maps generated from classification networks can indeed be class specific, as widely observed in the computer vision literature. Nevertheless, to achieve this, multiple classes are not typically present across all the images, which helps to find class discriminant regions (i.e., some images may contain several classes but the common situation is one class per image). In contrast, the different classes on the scans provided within the BraTS dataset often appear together, which results in finding specific class discriminant regions much harder. This hampers the performance of techniques designed to find class relevant regions, making the class activation maps sometimes useless. This observation has been already reported, for example in \cite{wu2019weakly}, who adopted a similar solution by considering only the whole tumor as the target class.}

The dataset is finally divided into training, validation and testing set, each containing 271, 32 and 32 scans, respectively and equally stratified with respect to high-grade gliomas (HGG) and low-grade gliomas (LGG) scans in each data split. \textcolor{black}{Slices within 3D-MRI scans were considered as 2D images, which with a dimension of 240 $\times$ 240. Last, to train the classification network that will generate the initial CAMs, images containing any tumor type were considered as \textit{non-healthy}, whereas the rest were identified as \textit{healthy}. Thus, in total we had 17,953 \textit{healthy} slices and 19,442 \textit{non-healthy} slices, excluding the blank slices with no skull-stripped region.} 

\paragraph{\textcolor{black}{\textbf{Prostate segmentation}}} We used the multi-modal prostate dataset from the popular DECATHLON segmentation challenge \cite{simpson2019large,antonelli2021medical} to localize Prostate central gland and peripheral zone. In particular, this dataset is composed of 32 volumetric scans containing MRI-T2 and apparent diffusion coefficient (ADC) maps, with their corresponding segmentation masks. Similar to the BraTS dataset, both the central gland and peripheral zone are combined into a single label for validation of the segmentation performance. The dataset is divided into training and validation splits, each containing 24 and 8 3D scans, respectively, and where each 2D slice in the scans was resized to 256 $\times$ 256 pixels. Due to the small size of the dataset, we performed a three-fold cross-validation and reported average results over the three runs.

\subsection{Evaluation metrics}

To assess the performance of the proposed approach, we employ the common Dice Similarity coefficient (DSC) and the average symmetric surface distance (ASSD). \textcolor{black}{As previously detailed, the ground truth segmentation is composed by the three (BraTS) and two (Prostate) classes, which are merged into a common mask.}

\subsection{Implementation details}

All models were implemented in PyTorch \cite{Paszke2019PyTorchAI} and use U-Net \cite{ronneberger2015u} as a backbone architecture. To train the models, we use AdamW \cite{loshchilov2019decoupled} optimizer with a learning rate of $5 \times 10^{-5}$, a weight decay of $0.1$ and a mini-batch size equal to 16. The regularization weight $\lambda_{KD}$ was empirically set to $0.5$ and  $\lambda_{E}(t)=\begin{cases}
e^{-(t-\mathcal{T})^{2}} & 0\leq t<\mathcal{T}\\
1 & \mathcal{T}\leq t
\end{cases}$ as a function of epochs $t$ with $\mathcal{T}=15$. We empirically observed that assigning large weights to the equivariance constraint terms at the beginning of the training hampered the quality of the class-activation maps. Thus, we employed a variable weighting coefficient that allows classification losses guiding the training to identify the initial CAMs, while equivariance losses gradually become important to revise the object regions across modalities. 

The affine image transformations include flipping, rotation, scaling, and translation. More concretely, the rotation transformation involves each image being randomly rotated with an angle equal to $(k\times90)^o$, where $k\in\{1,2,3\}$. During scaling, the scaling ratio is selected randomly in the range $[0.8,1.2]$. And last, the shifting in the translation transformation is selected randomly in the range $-0.3*(h,w)<(dh,dw)<0.3*(h,w)$, where $h$ and $w$ denote the height and width dimension of the given image. To demonstrate that the proposed learning strategy is robust against the CAM approach selected, we report the results on both CAMs \cite{zhou2016learning} and GradCAMs++ \cite{chattopadhay2018grad}. To obtain the final segmentation masks, we binarize all the CAMs with a threshold fixed to 0.5.

In our experiments we demonstrate the effectiveness of our approach by combining 2 and 4 modalities. In the two-modality scenario, we have created image pairs according to their \textit{a priori} dissimilarities. More concretely, T1 and T1c are the same modality with the only difference that T1c contains a contrast agent to improve the contrast of the boundaries in regions affected by hemorrhage. On the other hand,  FLAIR sequence is similar to T2, except that TE and TR times are much longer. Thus, we consider that T1 and T1c belong to one group, whereas T2 and FLAIR are included in another group. To create the image-pairs, we just combine one modality from the first group with one modality from the second group. Last, we demonstrate that our learning strategy is model-agnostic by reporting results on DeepLabv3.

Experiments were run in a server with 4 Nvidia P100 GPU cards. The code and trained models are made publicly available at: \url{https://github.com/gaurav104/WSS-CMER}

\begin{table*}[h!]
\caption{Quantitative results compared to prior literature. Individual performance represented in the first 2 columns, whereas combined results of individual maps are depicted in the last column. Best results are highlighted in bold.}
\centering
\tiny
\begin{tabular}{ll|cccc|cc}
\toprule
 & \textbf{}   & \multicolumn{2}{c}{\textbf{T1}} & \multicolumn{2}{c}{\textbf{T2}} & \multicolumn{2}{c}{\textbf{T1-T2}} \\
 \hline
 & \textbf{Method}       & \textbf{DSC}   & \textbf{ASSD}   & \textbf{DSC}   & \textbf{ASSD}   & \textbf{DSC}     & \textbf{ASSD}     \\
 \hline
\multirow{4}{*}{CAM}   & Baseline     &  38.66$\pm$16.27 & 12.25$\pm$7.08 &  46.80$\pm$15.90 & 10.83$\pm$9.05 &  49.81$\pm$14.78  & 11.04$\pm$5.73   \\
 & SEAM \cite{wang2020self} (scale) & 41.20$\pm$14.74 &  11.86$\pm$6.00  &  50.96$\pm$15.19    & 9.58$\pm$5.19  & 52.74$\pm$13.50   &  11.16$\pm$5.23  \\
 & SEAM \cite{wang2020self} (all)  & 39.05$\pm$14.32 &  10.74$\pm$5.06  &  50.11$\pm$14.34    & 7.58$\pm$4.02  & 54.23$\pm$12.20   & 10.09$\pm$4.91    \\
  
  & Proposed  & \bf 47.11$\pm$14.40  & \bf 10.43$\pm$5.55  & \bf 58.98$\pm$12.87 &  \bf 7.44$\pm$4.35 &  \bf \textcolor{black}{59.72$\pm$11.77} & \bf \textcolor{black}{9.91$\pm$5.09}   \\
  \hline

\multirow{4}{*}{GradCAM++} & Baseline &  39.16$\pm$16.11  & 12.19$\pm$6.95  &  45.79$\pm$15.59 & 10.92$\pm$9.06  &  50.11$\pm$14.95  & 11.14$\pm$5.72 \\
 & SEAM \cite{wang2020self} (scale)  & 41.94$\pm$14.46 & 11.86$\pm$5.85  &  51.34$\pm$15.23 & 9.66$\pm$5.25 &  52.70$\pm$13.61  &  11.32$\pm$5.21  \\
 & SEAM \cite{wang2020self} (all)  & 39.08$\pm$15.27 & 12.20$\pm$5.45  &  50.56$\pm$14.14 & 7.60$\pm$4.06 &  54.43$\pm$12.13  & 10.20$\pm$4.96  \\
    & Proposed  &  \bf 47.05$\pm$14.42  &  \bf 10.57$\pm$5.57   & \bf 58.93$\pm$12.80 &  \bf 7.47$\pm$4.36  & \bf \textcolor{black}{59.46$\pm$11.90}  & \bf \textcolor{black}{10.04$\pm$5.10}    \\
  \hline
  & \textbf{}   & \multicolumn{2}{c}{\textbf{T1c}} & \multicolumn{2}{c}{\textbf{ FLAIR}} & \multicolumn{2}{c}{\textbf{T1c-FLAIR}} \\
 \hline
 &        & \textbf{DSC}   & \textbf{ASSD}   & \textbf{DSC}   & \textbf{ASSD}   & \textbf{DSC}     & \textbf{ASSD}     \\
 \hline

\multirow{4}{*}{CAM}   & Baseline     & 36.27$\pm$18.33 &   14.36$\pm$8.48   &  41.58$\pm$14.66  &  7.87$\pm$4.33    &   45.40$\pm$14.78   &  11.06$\pm$5.34  \\
% & Baseline (2 Mod) &    &    &     &     &    &   \\

  & SEAM \cite{wang2020self} (scale)  &  40.91$\pm$17.20  &  12.16$\pm$4.77  &  41.32$\pm$15.11   & \bf 7.23$\pm$3.84  &  48.01$\pm$12.81   &  9.62$\pm$4.45  \\

  & SEAM \cite{wang2020self} (all) &  43.38$\pm$16.79 & 12.40$\pm$6.52  &  43.04$\pm$14.40   & 7.77$\pm$4.67  & 51.31$\pm$12.44   &  9.61$\pm$5.01  \\
    & Proposed   &   \bf 47.95$\pm$15.43  &  \bf 11.50$\pm$6.54   & \bf 53.05$\pm$16.52 &  8.34$\pm$4.44   &  \bf \textcolor{black}{57.59$\pm$15.00}   &  \bf \textcolor{black}{9.43$\pm$5.09}   \\
  \hline
\multirow{4}{*}{GradCAM++} & Baseline      &      37.23$\pm$18.60          &        14.35$\pm$8.45        &        43.36$\pm$14.42        &       7.89$\pm$4.39         &          46.32$\pm$14.77        &      11.21$\pm$5.38            \\
% & Baseline (2 Mod) &    &    &     &     &    &   \\

  & SEAM \cite{wang2020self} (scale)  &  41.54$\pm$17.44  &  12.18$\pm$5.08  &  42.56$\pm$14.84   & \bf 7.23$\pm$3.84  &  47.96$\pm$13.52  &  9.67$\pm$4.48  \\
  & SEAM \cite{wang2020self} (all)  & 43.45$\pm$16.86 & 12.45$\pm$6.49  &  43.82$\pm$14.44   & 7.82$\pm$4.63  &  51.58$\pm$12.43  &  9.72$\pm$5.03  \\
   & Proposed   &   \bf 47.88$\pm$15.41  &  \bf 11.57$\pm$6.53   & \bf 53.02$\pm$16.46 &  8.37$\pm$4.46  & \bf \textcolor{black}{57.55$\pm$15.00}  & \bf \textcolor{black}{9.44$\pm$5.09}  \\ 
    \hline
  & \textbf{}   & \multicolumn{2}{c}{\textbf{T1}} & \multicolumn{2}{c}{\textbf{ FLAIR}} & \multicolumn{2}{c}{\textbf{T1-FLAIR}} \\
 \hline
 &        & \textbf{DSC}   & \textbf{ASSD}   & \textbf{DSC}   & \textbf{ASSD}   & \textbf{DSC}     & \textbf{ASSD}     \\
 \hline
\multirow{4}{*}{CAM}   & Baseline     & 38.66$\pm$16.27 & 12.25$\pm$7.08   &  41.58$\pm$14.66  &  7.87$\pm$4.33   &   48.42$\pm$13.53   &  11.04$\pm$5.73  \\
  & SEAM \cite{wang2020self} (scale)  &  41.20$\pm$14.74 &  11.86$\pm$6.00 &  41.32$\pm$15.11   & \bf 7.23$\pm$3.84 &  48.68$\pm$13.06  &  9.60$\pm$4.98  \\
  & SEAM \cite{wang2020self} (all)  &  39.05$\pm$14.32 &  \bf 10.74$\pm$5.06  &  43.04$\pm$14.40   & 7.77$\pm$4.67 &  49.82$\pm$13.00  &  9.15$\pm$4.70  \\
    & Proposed   &  \bf 47.23$\pm$13.19  &  11.12$\pm$4.92   & \bf 56.20$\pm$18.76 &  7.50$\pm$4.20 &  \bf \textcolor{black}{58.87$\pm$15.04}  & \bf \textcolor{black}{9.01$\pm$4.66}   \\
  \hline
\multirow{4}{*}{GradCAM++} & Baseline     &    39.16$\pm$16.11  & 12.19$\pm$6.95    &         43.36$\pm$14.42        &       7.89$\pm$4.39           &         49.23$\pm$13.54         &  11.13$\pm$5.72                \\
  & SEAM \cite{wang2020self} (scale)  &  41.94$\pm$14.46 & 11.86$\pm$5.85  &    42.56$\pm$14.84   & \bf 7.23$\pm$3.84  &  49.13$\pm$13.09  &  9.78$\pm$4.99  \\
  & SEAM \cite{wang2020self} (all)  &  39.08$\pm$15.27 & 12.20$\pm$5.45  &  43.82$\pm$14.44   & 7.82$\pm$4.63  &  50.44$\pm$12.95  &   9.38$\pm$4.85 \\
    & Proposed   &  \bf 47.30$\pm$12.70  &  \bf 11.21$\pm$4.97   & \bf 56.34$\pm$18.50 &  7.59$\pm$4.30 &  \bf \textcolor{black}{58.53$\pm$15.09}  &   \bf  \textcolor{black}{9.13$\pm$4.66} \\ 
    \hline
  & \textbf{}   & \multicolumn{2}{c}{\textbf{T1c}} & \multicolumn{2}{c}{\textbf{T2}} & \multicolumn{2}{c}{\textbf{T1c-T2}} \\
 \hline
 &        & \textbf{DSC}   & \textbf{ASSD}   & \textbf{DSC}   & \textbf{ASSD}   & \textbf{DSC}     & \textbf{ASSD}     \\
 \hline
\multirow{4}{*}{CAM}   & Baseline    & 36.27$\pm$18.33 &   14.36$\pm$8.48  &  46.80$\pm$15.90 & 10.83$\pm$9.05  &   47.44$\pm$15.27   &  12.56$\pm$5.90  \\
% & Baseline (2 Mod) &    &    &     &     &    &   \\
  & SEAM \cite{wang2020self} (scale)  &  40.91$\pm$17.20  &  12.16$\pm$4.77  &  50.96$\pm$15.19    & 9.58$\pm$5.19 &  51.53$\pm$12.82  &  10.82$\pm$4.69  \\
  & SEAM \cite{wang2020self} (all)  & 43.38$\pm$16.79 & 12.40$\pm$6.52  &  50.11$\pm$14.34    & \bf 7.58$\pm$4.02 &  56.01$\pm$11.96  & 10.53$\pm$5.01 \\
    & Proposed  & \bf 47.47$\pm$15.05  &  \bf 11.85$\pm$5.66  & \bf 57.90$\pm$13.85 &  8.39$\pm$5.40  &  \bf \textcolor{black}{59.16$\pm$11.33}  &  \bf \textcolor{black}{10.35$\pm$5.16}   \\
  \hline
\multirow{4}{*}{GradCAM++} & Baseline     &     37.23$\pm$18.60          &        14.35$\pm$8.45    &           45.79$\pm$15.59 & 10.92$\pm$9.06  & 47.82$\pm$15.59 & 12.67$\pm$5.92                 \\
% & Baseline (2 Mod) &    &    &     &     &    &   \\
  & SEAM \cite{wang2020self} (scale)  &  41.54$\pm$17.44  &  12.18$\pm$5.08  &  51.34$\pm$15.23 & 9.66$\pm$5.25 &  51.73$\pm$12.94  &   10.85$\pm$4.72 \\
  & SEAM \cite{wang2020self} (all)  & 43.45$\pm$16.86 & 12.45$\pm$6.49  & 50.56$\pm$14.14 & \bf 7.60$\pm$4.06 &  56.14$\pm$11.93  &  10.56$\pm$5.02  \\
    & Proposed   &  \bf 47.59$\pm$15.02  &  \bf 11.92$\pm$5.62   & \bf 57.97$\pm$13.59 &  8.57$\pm$5.58  &  \bf \textcolor{black}{58.93$\pm$13.41}   &  \bf \textcolor{black}{10.44$\pm$5.16}   \\ \bottomrule     
\end{tabular}
\label{table:main}
\end{table*}

\subsection{Baselines for comparison} We benchmark the proposed approach to relevant prior literature. As our approach resorts to CAM and GradCAM++ to generate the final pixel-wise segmentations, we employ them as lower bound baselines. Then, we also include SEAM \cite{wang2020self} in our experiments. The reason behind this choice is two-fold. First, SEAM also incorporates equivariance regularizers to boost the initial CAMs. In this case, while in the original work authors report results on scale equivariance, we also evaluate their performance when including several image transformations. And second, as reported in their experiments, this method represents the current state-of-the-art in weakly supervised segmentation. Last, we also include the results from a fully-supervised model trained with a combination of the dice and the cross-entropy loss as the objective function, which represents the upper bound. In particular, to assess the impact of the proposed learning objectives, the upper bound network is the same architecture used as backbone in the proposed pipeline with the only difference that it is trained on pixel-wise annotations. We would like to emphasize that the goal of this work is not to obtain new state-of-the-art results on the segmentation task, but to demonstrate that our method can approach the gap between weakly and fully supervised segmentation models. This motivates our choice of employing a standard UNet in our experiments.

\section{Results}

\subsection{Quantitative results}

\begin{table*}[t!]
\caption{Quantitative results compared to prior literature. Individual performance represented in the first 4 columns, whereas combined results of individual maps are depicted in the last column. Best results are highlighted in bold.}
\tiny
\centering
\begin{tabular}{ll|cccccccc|cc}
\hline
                           &                                                   & \multicolumn{2}{c}{\textbf{T1}}  & \multicolumn{2}{c}{\textbf{T2}}  & \multicolumn{2}{c}{\textbf{T1c}} & \multicolumn{2}{c}{\textbf{ FLAIR}} & \multicolumn{2}{c}{\textbf{Fused}} \\ \hline
                           & \multicolumn{1}{c|}{\textbf{Method}}              & \textbf{DSC}    & \textbf{ASSD}   & \textbf{DSC}    & \textbf{ASSD}   & \textbf{DSC}    & \textbf{ASSD}   & \textbf{DSC}      & \textbf{ASSD}   & \textbf{DSC}      & \textbf{ASSD}    \\ \hline
\multirow{4}{*}{CAM}       & Baseline                                          & 38.66$\pm$16.27 & 12.25$\pm$7.08 & 46.80$\pm$15.90 & 10.83$\pm$9.05 & 36.27$\pm$18.33 & 14.36$\pm$8.48 & 41.58$\pm$14.66   & 7.87$\pm$4.33  & 49.64$\pm$14.05   & 11.28$\pm$5.46  \\
                           & SEAM\cite{wang2020self} (scale)  & 41.20$\pm$14.74 & 11.86$\pm$6.00 & 50.96$\pm$15.19 & 9.58$\pm$5.19  & 40.91$\pm$17.20 & 12.16$\pm$4.77 & 41.32$\pm$15.11   & \bf 7.23$\pm$3.84  & 51.72$\pm$13.39   & 11.13$\pm$4.89  \\
                           & SEAM \cite{wang2020self} (all)   & 39.05$\pm$14.32 & 10.74$\pm$5.06 & 50.11$\pm$14.34    & \bf 7.58$\pm$4.02  & 43.38$\pm$16.79 & 12.40$\pm$6.52 & 43.04$\pm$14.40   & 7.77$\pm$4.67  & 54.53$\pm$11.14   & 11.00$\pm$5.00  \\
                            & Proposed                                          & \bf 47.56$\pm$14.81 & \bf 10.31$\pm$5.01 & \bf 57.07$\pm$14.23 & 8.39$\pm$5.09  & \bf 44.74$\pm$15.99  &   \bf 11.62$\pm$6.12 & \bf 58.64$\pm$16.42   & 7.29$\pm$4.38  & \bf \textcolor{black}{56.81$\pm$13.84}   & \bf \textcolor{black}{10.80$\pm$5.01 }   \\ \hline
\multirow{4}{*}{GradCAM++} & Baseline                                          & 39.16$\pm$16.11 & 12.19$\pm$6.95 & 45.79$\pm$15.59 & 10.92$\pm$9.06 & 37.23$\pm$18.60 & 14.35$\pm$8.45 & 43.36$\pm$14.42   & 7.89$\pm$4.39  & 49.98$\pm$14.22   & 11.41$\pm$5.47  \\
                           & SEAM \cite{wang2020self} (scale) & 41.94$\pm$14.46 & 11.86$\pm$5.85 & 51.34$\pm$15.23 & 9.66$\pm$5.25  & 41.54$\pm$17.44 & 12.18$\pm$5.08 & 42.56$\pm$14.84   & \bf 7.23$\pm$3.84  & 51.78$\pm$13.51   & 11.24$\pm$4.92  \\
                           & SEAM \cite{wang2020self} (all)   & 39.08$\pm$15.27 & 12.20$\pm$5.45 & 50.56$\pm$14.14 & \bf 7.60$\pm$4.06  & 43.45$\pm$16.86 & 12.45$\pm$6.49 & 43.82$\pm$14.44   & 7.82$\pm$4.63  & 54.64$\pm$11.17   & 11.07$\pm$5.02  \\
                            & Proposed                                          & \bf 47.48$\pm$14.35 & \bf 10.39$\pm$5.00 &\bf  57.13$\pm$14.17 & 8.42$\pm$5.12  & \bf 44.73$\pm$15.99  &  \bf 11.62$\pm$6.18 & \bf 59.73$\pm$16.41   & 7.26$\pm$4.49  & \bf \textcolor{black}{57.05$\pm$14.06}   &
                         \bf  \textcolor{black}{10.83$\pm$5.01}  \\ \hline
Upper Bound & Full Supervision & 71.40$\pm$17.15 &  6.37$\pm$3.71 &  79.60$\pm$14.88 &  3.04$\pm$1.87 &  66.54$\pm$24.42 &  5.27$\pm$4.26 &  81.80$\pm$17.10 &  2.30$\pm$1.50 & -- & --\\
\hline
\end{tabular}
\label{table:main_4mod}
\end{table*}

\begin{table*}[]
\caption{Ablation study on the losses: 
$\mathcal{L}_{KD}$ (Eq. \ref{eq:kl}), $\mathcal{L}_{ER}$ (Eq. \ref{eq:ER}) and $\mathcal{L}_{CMER}$ (Eq. \ref{eq:cmer}). Best results in bold.}
\tiny
\centering
\begin{tabular}{lcccllllllllll}
\toprule
                            & \multicolumn{3}{c}{\textbf{Losses}}                            & \multicolumn{2}{c}{\textbf{T1-T2}}                                     & \multicolumn{2}{c}{\textbf{T1c- FLAIR}}                                     & \multicolumn{2}{c}{\textbf{T1- FLAIR}}                                    & \multicolumn{2}{c}{\textbf{T1c-T2}}                                  & \multicolumn{2}{c}{\textbf{All}}                                    \\ \midrule
                            & $\mathcal{L}_{KD}$ & $\mathcal{L}_{ER}$ & $\mathcal{L}_{CMER}$ & \multicolumn{1}{c}{\textbf{DSC}} & \multicolumn{1}{c}{\textbf{ASSD}} & \multicolumn{1}{c}{\textbf{DSC}} & \multicolumn{1}{c}{\textbf{ASSD}} & \multicolumn{1}{c}{\textbf{DSC}} & \multicolumn{1}{c}{\textbf{ASSD}} & \multicolumn{1}{c}{\textbf{DSC}} & \multicolumn{1}{c}{\textbf{ASSD}} & \multicolumn{1}{c}{\textbf{DSC}} & \multicolumn{1}{c}{\textbf{ASSD}} \\ \midrule
\multirow{4}{*}{CAM}        & $\checkmark$       & --                 & --                   & 55.91$\pm$13.93     &   10.46$\pm$4.68    &  51.39$\pm$13.46    &    11.92$\pm$4.85           &     50.08$\pm$13.51    &  10.87$\pm$4.72   & 51.39$\pm$13.46    &    11.92$\pm$4.85           & 46.44$\pm$13.98                  & 12.07$\pm$4.80                 \\
                            & $\checkmark$                & $\checkmark$       & --        & 55.13$\pm$13.21  & 12.18$\pm$5.61            & 55.46$\pm$13.37  & 10.81$\pm$5.28          &  52.94$\pm$14.17  & 11.44$\pm$5.28            &                 55.46$\pm$13.37  & 10.81$\pm$5.28           & 54.37$\pm$16.20                  & 12.98$\pm$5.63                         \\
                            & --       & $\checkmark$       & $\checkmark$                   & \bf 60.39$\pm$11.60  & \bf 9.19$\pm$4.76                &  57.07$\pm$11.92  & 11.23$\pm$5.19   &  57.24$\pm$14.57  &  9.04$\pm$4.59   &       57.07$\pm$11.92  & 11.23$\pm$5.19    & 55.62$\pm$14.26                  & 11.10$\pm$4.89                           \\
                            & $\checkmark$       & $\checkmark$       & $\checkmark$         &  \textcolor{black}{59.72$\pm$11.77} & \textcolor{black}{9.91$\pm$5.09}   &   \bf \textcolor{black}{57.59$\pm$15.00}   &  \bf \textcolor{black}{9.43$\pm$5.09}   &  \bf \textcolor{black}{58.87$\pm$15.04}  & \bf \textcolor{black}{9.01$\pm$4.66} &  \bf \textcolor{black}{59.16$\pm$13.33}  & \bf \textcolor{black}{10.35$\pm$5.16}   &  \bf \textcolor{black}{56.81$\pm$13.84}     &     \bf \textcolor{black}{10.80$\pm$5.01 }             \\
                            \midrule
\multirow{4}{*}{GradCAM++}        & $\checkmark$       & --                 & --                   & 55.88$\pm$13.87  &  10.58$\pm$4.70  &  51.15$\pm$13.53  &  12.12$\pm$4.85  & 50.28$\pm$13.76  &  11.13$\pm$4.83 & 51.15$\pm$13.53 &  12.12$\pm$4.85  & 46.15$\pm$14.04                  & 12.45$\pm$4.81                  \\
                            & $\checkmark$                & $\checkmark$       & --        & 54.15$\pm$13.60  & 12.67$\pm$5.79    & 55.22$\pm$13.48  & 10.84$\pm$5.29 & 53.03$\pm$14.25  & 11.61$\pm$5.35  &                55.22$\pm$13.48  & 10.84$\pm$5.29     &    54.29$\pm$16.21                  & 13.09$\pm$5.63               \\
                            & --       & $\checkmark$       & $\checkmark$                   & \bf 60.26$\pm$11.62  & \bf 9.28$\pm$4.81     &  56.80$\pm$12.01  & 11.38$\pm$5.18 &  56.99$\pm$14.62  & 9.30$\pm$4.76  &  56.80$\pm$12.01  & 11.38$\pm$5.18  &  55.43$\pm$14.28                  & 11.2$\pm$4.89    \\
                            & $\checkmark$       & $\checkmark$       & $\checkmark$         &  \textcolor{black}{59.46$\pm$11.90} & \textcolor{black}{10.04$\pm$5.10}  &   \bf \textcolor{black}{57.55$\pm$15.00}  & \bf \textcolor{black}{9.44$\pm$5.09} &  
                            \bf \textcolor{black}{58.53$\pm$15.09}  &  \bf \textcolor{black}{9.13$\pm$4.66} &  \bf \textcolor{black}{58.93$\pm$13.41}   &  \bf \textcolor{black}{10.44$\pm$5.16}   &   \bf \textcolor{black}{57.05$\pm$14.06}   &        \bf \textcolor{black}{10.83$\pm$5.01}         \\                   
                        
 \bottomrule
\end{tabular}
\label{table:ablationLoss}
\end{table*}

\subsubsection{Comparison to the literature}

The results of this experiment are reported in Table \ref{table:main}, where the first two columns show the individual results per modality and the final column contains the results when both modalities are combined. CAMs fusion is achieved by getting the maximum pixel-wise value across modalities, i.e., $\max(\mathbf{M}^1,\mathbf{M}^2)$. If we look at individual-modality performances, the first thing that we can observe is that the original SEAM model, which only employs the scale transformation as equivariance regularizer, consistently improves the baselines across all modalities. Interestingly, the fact of including additional image transformations (i.e., SEAM-all) does not necessarily improve the performance, contrary to the common belief followed in standard data-augmentation strategies. Indeed, this finding aligns with the observations in \cite{wang2020self}, where authors found that simply aggregating different affine transformations does not always bring segmentation improvements. 

In our experiments, we demonstrate that certain modalities (i.e., T1c and  FLAIR) benefit from multiple image affine transformations, whereas others (i.e., T1 and T2) see their performance degrade. Regarding the proposed method, we observe a significant improvement over the baselines, which is consistent across modalities and CAM versions. In particular, compared to the original SEAM approach \cite{wang2020self}, the proposed model brings between 6\% (in T1,T1c) to 12\% (in  FLAIR) improvement in terms of DSC. Differences in the ASSD metric are smaller, with values ranging from 0.3 to 1 voxels, as average, for all the modalities except  FLAIR, where SEAM slightly outperforms the proposed model. Note that our method employs the same affine transformations as SEAM-all and therefore, performance differences are further magnified with respect to this model. Last, we can observe that this trend holds when fusing both modalities (\textit{last column}).  In particular, our model outperforms the second best model between nearly 3\% (T1-T2) to more than 8\% (T1- FLAIR).  We also report the values obtained on the four modality scenario (Table \ref{table:main_4mod}).  Similar to the two-modalities setting, our model substantially improves the performance over the other models. Including multi-modal information flow during training results in performance gains between 7\% and 16\% in terms of DSC, such as in T1, T2 and  FLAIR. If individual modality results are further combined, improvement over the baseline is around 7\% compared to the original CAMs and GradCAM++. Last, our model outperforms the two settings based in \cite{wang2020self} by 5\% and nearly 2.5\%. These results suggest that by facilitating information flow between modalities by means of inter-modality equivariance constraints on original CAMs significantly improves the segmentation performance.

We also report the results of the upper bound in Table \ref{table:main_4mod} across individual modalities. These values show that despite our method obtains closer results to standard fully supervised learning, there are still opportunities to progress in this task. We would like to emphasize that the upper bound employed in this work is a standard U-Net, which explains the differences with respect to the top-ranked approaches in the leaderboard of the BraTS Challenge. The goal of these works is to boost the performance of fully supervised models, which is typically achieved by integrating more sophisticated modules or network architectures. On the other hand, our objective is to show that with the same model as backbone, our learning strategy can make better use of multi-modal images, achieving results closer to those obtained by the fully supervised upper bound.

\subsubsection{On the impact of the different objective terms}

We now assess the effect of the learning objectives integrated in our model. To this end, we report the results of four image-pairs, as well as a model containing the four modalities (Table \ref{table:ablationLoss}). These results show that including the within-modality equivariance constraint (Eq. \ref{eq:ER}) typically results in an improvement that ranges across settings. For instance, in terms of DSC, the performance gain in the tuple T1c- FLAIR is nearly 3\%, whereas the improvement in the all-modalities scenario is close to 8\%. If we also include the cross-modality equivariance term during training (Eq. \ref{eq:cmer}) the performance is further increased, with values ranging from 3\% to 4.5\%. 

\subsubsection{Ablation study on equivariance}
\begin{figure}[h!]
        \centering
     \includegraphics[width=\linewidth]{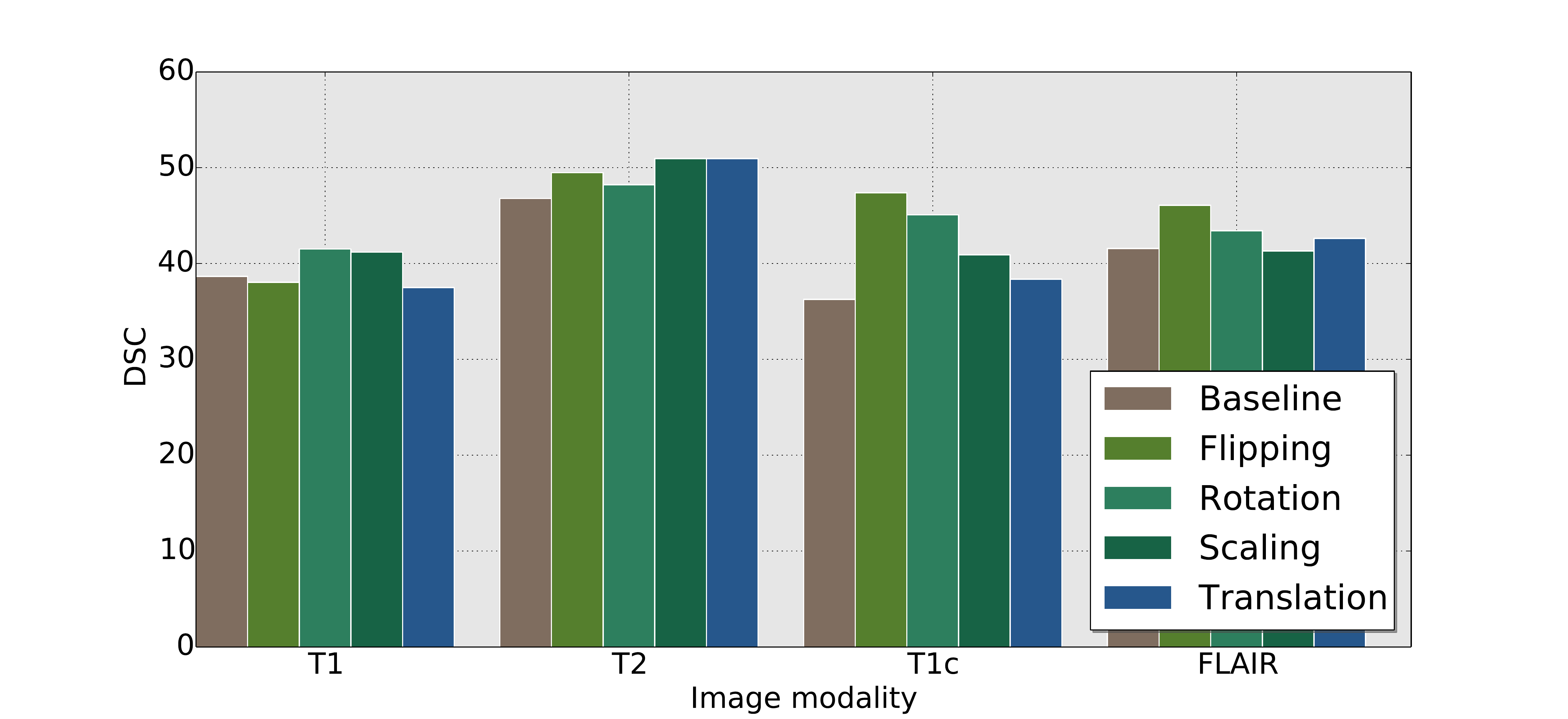}
          
        \caption{Impact on predictions when several transformations are employed in the equivariant constraints.}
        
        \label{fig:equivariant}
    \end{figure}

Theoretically, we can employ any spatial transformation $\pi(\cdot)$ to explicitly enforce equivariant constraints. Nevertheless, in practice, not all the affine transformations impact each modality equally. Thus, we assess the effect of several affine transformations in this ablation study, whose results are depicted in Fig \ref{fig:equivariant}. We can observe several interesting observations. First, there exist transformations that always improve the performance over the baseline (e.g., rotation), whereas others either enhance or degrade the results depending on the target modality (e.g., flipping, scaling and translation). Furthermore, this negative impact is not consistent across modalities. For example, using translation as equivariance constraint degrades the performance on T1 and while has a positive effect on T2, T1c and  FLAIR. On the other hand, scaling the images boosts the performance on T1, T1c and T2, but it has a degrading effect on  FLAIR. Thus, finding a set of ideal affine transformations that work well across modalities is not straightforward, as demonstrated in this experiment. This motivates us to employ jointly all the transformations.

\subsubsection{Robustness to backbone}

In this section we evaluate whether changes in the segmentation backbone impact the trend observed in the main results. In Table \ref{table:ablat_deeplab}, it can observed that if we replace UNet by DeepLabv3\cite{chen2017rethinking} with ResNet-50 as backbone, the gap between the proposed method and prior literature still holds, being larger in some cases. For example, in the T1-T2 and T1c-FLAIR pairs, differences between our approach and SEAM \cite{wang2020self} (all) range from 6.5\% to 12\% with DeepLabv3 as backbone, where the difference was of 5\% and 6\%, respectively, with UNet. This suggests that our learning strategy is model-agnostic and can be employed with any segmentation network.

\begin{table}[h!]
\caption{Results, in terms of DSC, with DeepLabv3 as backbone architecture. Best results are highlighted in bold.}
\tiny
%\centering{}{\scriptsize{}}%
\centering
\begin{tabular}{>{\raggedright}m{0.05\textwidth}lccccc}
\toprule
 & \textbf{Method} & \textbf{T1-T2} & \textbf{T1c-FLAIR} & \textbf{T1-FLAIR} & \textbf{T1c-T2}\tabularnewline
\midrule
\multirow{4}{0.1\textwidth}{ CAM} & Baseline & 50.07$\pm$13.46 &  48.26$\pm$14.73 & 49.42$\pm$12.69 &  48.72$\pm$15.39 \tabularnewline
 & SEAM (scale) & 52.01$\pm$14.21 & 51.43$\pm$15.27 & 54.27$\pm$13.58 & 52.60$\pm$13.23 \tabularnewline
  & SEAM (all) & 54.38$\pm$14.04 & 51.09$\pm$12.18 & 54.21$\pm$11.49 & 57.65$\pm$14.73 \tabularnewline
   & Proposed & \bf 60.78$\pm$11.05 & \bf 62.32$\pm$14.23 & \bf 62.34$\pm$11.88 & \bf 60.92$\pm$13.38 \tabularnewline
\midrule
\multirow{4}{0.1\textwidth}{GradCAM++} & Baseline & 49.98$\pm$13.47 &  48.16$\pm$14.94  & 49.49$\pm$12.86  &  48.42$\pm$15.38 \tabularnewline
 & SEAM (scale) & 52.07$\pm$14.19 & 50.34$\pm$15.45 & 53.62$\pm$13.76 & 52.62$\pm$13.24 \tabularnewline
  & SEAM (all) & 54.20$\pm$14.12 & 50.47$\pm$13.11 & 54.13$\pm$11.58 & 56.55$\pm$15.82 \tabularnewline
    & Proposed & \bf 60.95$\pm$10.97 & \bf 62.40$\pm$14.32 & \bf 60.08$\pm$11.85 & \bf 60.20$\pm$14.30 \tabularnewline
\bottomrule
\end{tabular}
\label{table:ablat_deeplab} %ablat_backbone
{\scriptsize\par}
\end{table}

\subsubsection{Source of improvement}
\begin{figure}[h!]
\centering
\includegraphics[width=1\linewidth]{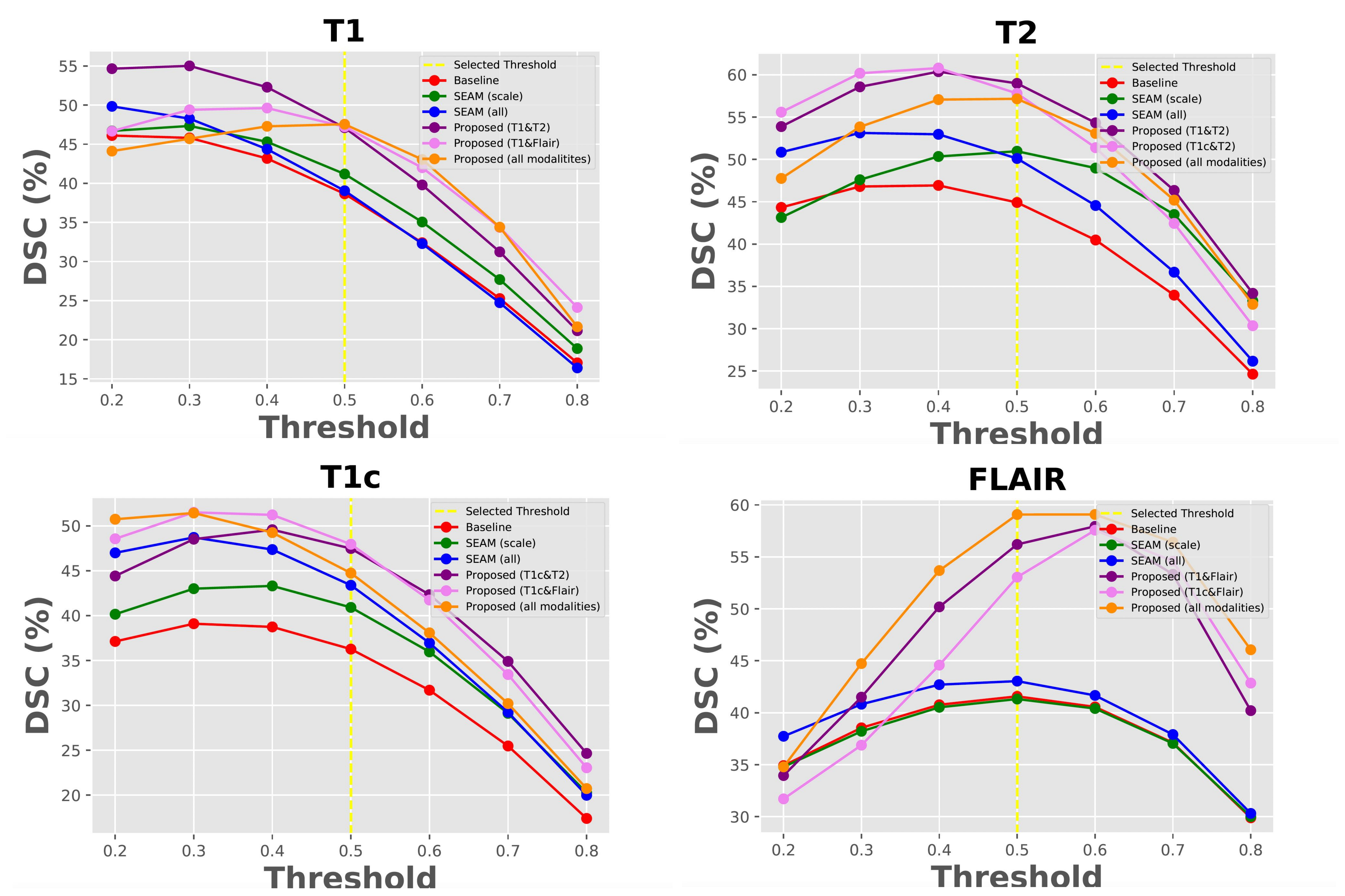}
\caption{Variation of DSC with respect to the threshold selected to generate the CAMs in each modality. We employ three lines to represent our methods: two for the two-modalities setting and one for the four-modality case. Note that the other approaches do not leverage multi-modal information.}
        
        \label{fig:variation}
\end{figure}

Qualitatively (e.g., Fig \ref{fig:Class-Activation-maps} and \ref{fig:visual-results}), we can observe that the improvement on CAMs arises from better coverage of activated regions and a reduction of over activated areas. Nevertheless, to further investigate the source of improvement, we depict in Figure \ref{fig:variation} the variation of the DSC across different threshold values selected to binarize the obtained CAMs for each modality. First, we can observe that the proposed approach generates CAMs which bring consistent improvements across the threshold values, compared to prior works. Then, for some modalities, e.g., T1, T2 or FLAIR our approach generates CAMs whose best thresholds are centered around 0.5. 
This suggests that cross-modality information flow might activate under-activated regions and suppress over-activations of the single-modality baselines, resulting in more consistent CAMs across all thresholds. 

To quantitatively validate this hypothesis, we compute the ratio between False Negatives (FN) and True Positives (TP), i.e.,  r$_u=FN/TP$ and the ratio between False Positives (FP) and TP,  i.e.,  r$_o=FP/TP$ with the threshold set to 0.5. Large number of FN will indicate potential under segmentations, whereas large number of FP represents over-segmented CAMs. Thus, the larger the values of these ratios the worst the generated CAMs. We report in Table \ref{table:ratios} the computed r$_u$ and r$_o$ ratios across the different modalities compared to the SEAM method. As expected, the proposed model generates the lowest scores for both ratios, which is consistent across modalities, indicating that our approach reduces over-activated pixels while covers more complete regions. This proves that the intra and cross-modality equivariant constraints proposed in this work positively contribute during learning, leading to noticeable CAM improvements.

\begin{table}[h!]
\caption{Ratios highlighting the source of improvement. Best results in bold.}
\centering
\scriptsize
\begin{tabular}{lc|ccc}
\hline
\bf Modality&  \bf Ratio & 
\bf SEAM (scale) & \bf SEAM (all) & \bf Proposed \\
\hline
\multirow{2}{0.01\textwidth}{\bf T1} & r$_u$ & 
3.65 & 3.55 &  \bf 2.33 \\
 & r$_o$ & 
 3.51 & 2.49 &  \bf 1.93\\
 \hline
 \multirow{2}{0.01\textwidth}{\bf T2} & r$_u$ &
 1.59 & 1.72 &  \bf 0.94 \\
 & r$_o$ &  
 2.07 & 1.38 &  \bf 0.76 \\
 \hline
 \multirow{2}{0.01\textwidth}{\bf T1c} & r$_u$ &
 2.64 & 2.58 &  \bf 2.42 \\
 & r$_o$ &% 
 2.40 & 2.47 &  \bf 1.89 \\
 \hline
 \multirow{2}{0.01\textwidth}{\bf FLAIR} & r$_u$ & 
 1.76 & 1.31 &  \bf 0.73 \\
 & r$_o$ & 
 2.17 & 2.54 &  \bf 0.81 \\
 \hline
\label{table:ratios}
\end{tabular}
\end{table}

\vspace{-1mm}
    
\subsection{Qualitative results}

\begin{figure}[h!]
\begin{centering}
\includegraphics[width=1\linewidth]{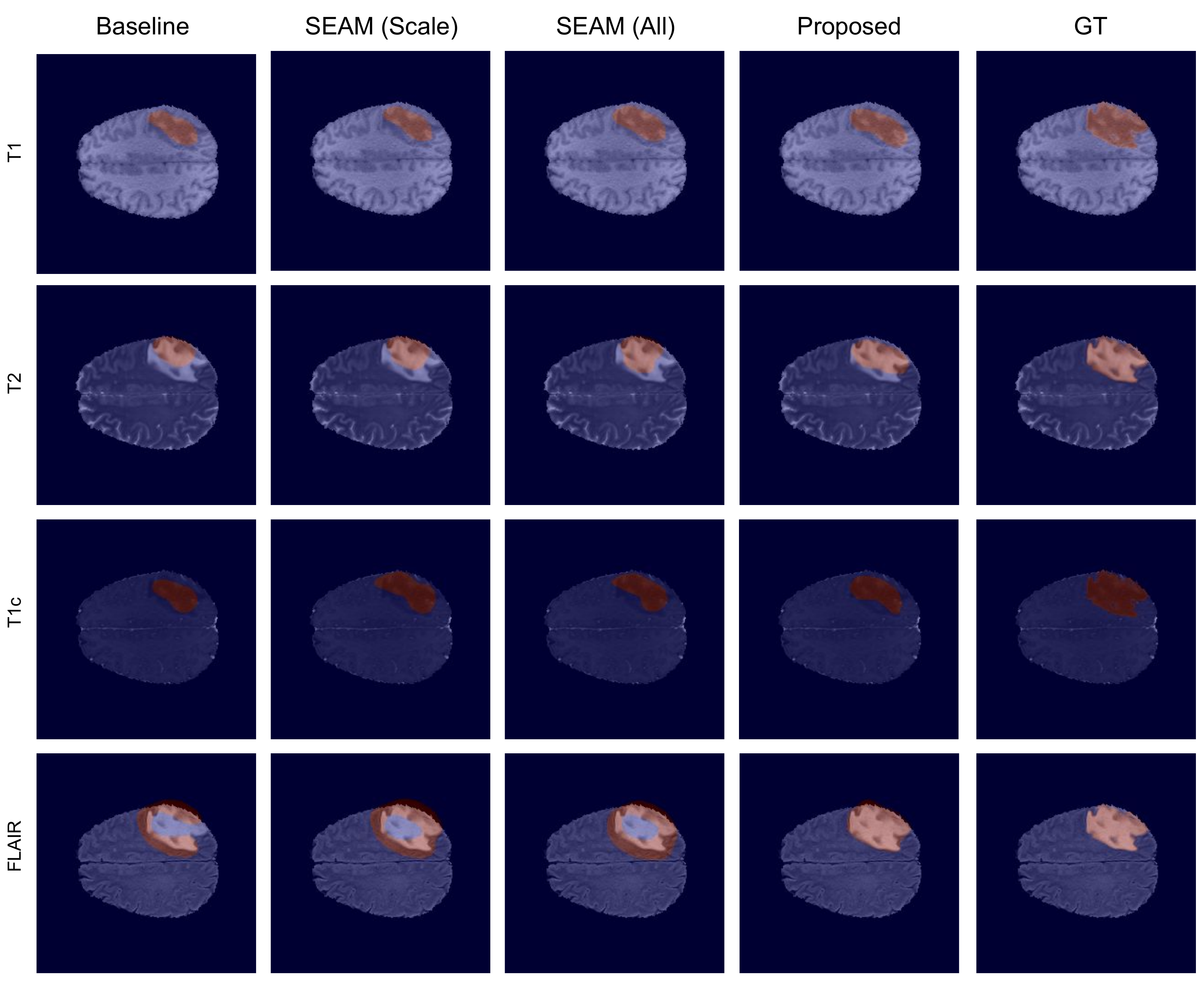}
\par\end{centering}
\caption{Qualitative results on a given test scan (the binarized CAM is overlaid on the original image). Rows represent the modality whereas the different models, as well as the corresponding ground truth are shown in columns.}
 
\label{fig:visual-results}
\end{figure}

Figure \ref{fig:visual-results} depicts the visual results across different models. We can observe that the original CAMs (\textit{first column}) typically results in undersegmented regions, which aligns with the observations in the literature. Integrating single-modality equivariant constraints (i.e, SEAM\cite{wang2020self}) has shown in improve the quantitative performance. Nevertheless, as we can observe in these images, this does not translate into an significant enhancement in terms of visual results. Indeed, SEAM based models slightly expand the initial CAMs across all the modalities, particularly in T1c and FLAIR. Nevertheless, the region coverage by this model still misses large lesion areas. Finally, it can be observed that the CAMs generated by our model better align with the ground truth. This improvement stems not only from more complete activated regions (in all the modalities) but also from a reduction in over-activated areas (in FLAIR).

\subsection{\textcolor{black}{Implicit data augmentation or explicit equivariant constraints?}}

\begin{figure}[h!]
\begin{centering}
\includegraphics[width=1\linewidth]{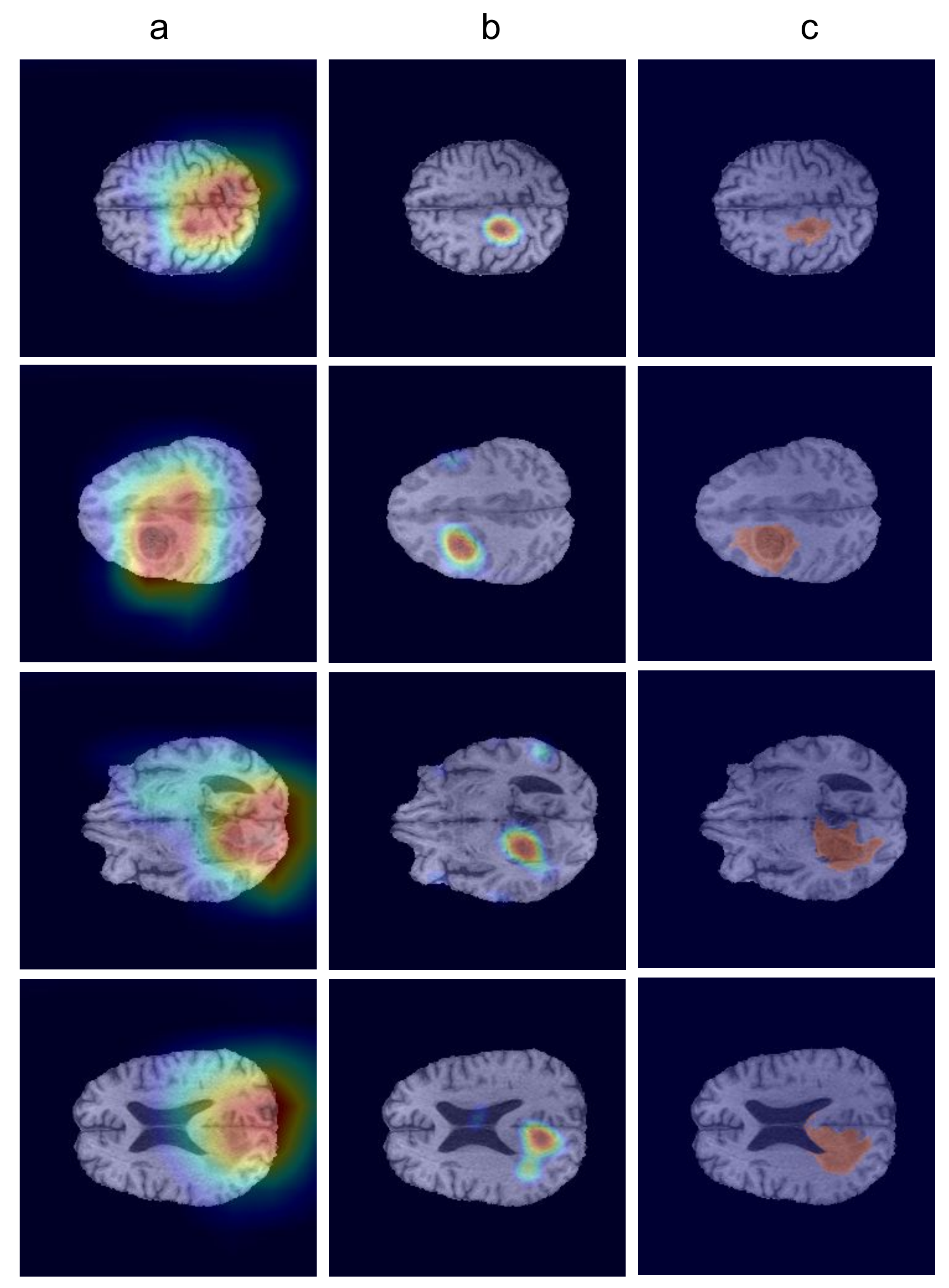}
\par\end{centering}
\caption{\textcolor{black}{Qualitative results to assess the impact on the performance of directly using data augmentation (\textit{a}) versus explicitly using the proposed equivariant constraints (\textit{b}). For comparison purposes, the segmentation ground truth for each slice is provided in (\textit{c}).}}
 
\label{fig:visual-results-equi}
\end{figure}

\textcolor{black}{To visually explain the benefits of integrating our equivariant constraints over the use of \textit{ad-hoc} data augmentation, we have depicted several class attention maps in Figure \ref{fig:visual-results-equi}. In particular, we can see that in Figure \ref{fig:visual-results-equi}(\textit{a}), which represents the model trained with the standard \textit{on-the-fly} augmentation approach, consisting of the same affine transforms employed for equivariance, the generated CAMs fail to identify distinctive tumor regions, resulting in substantially large over-segmentations. The poor quality of these CAMs can be easily explained. When data augmentation is performed, affine transformations attempt to implicitly introduce equivariant constraints on the output space. For example, if a rotated version of a given image is generated, the predicted segmentation of the transformed image should be the same as the rotated segmentation of the original image. Nevertheless, this implicit constraint is lost in the resulting CAMs, as they are obtained from the image level labels, resulting in large inconsistencies of highlighted regions among multiple versions of the same image. This behaviour has been also observed in \cite{wang2020self}, suggesting that the use of data augmentation does not necessarily improves the quality of generated CAMs. On the other hand, the proposed approach has the capability of explicitly enforcing equivariant CAMs, which are leveraged via the additional losses presented in Section \ref{ssec:self}. }

\subsection{\textcolor{black}{Generalization to other datasets}}

\begin{table*}[h!]
\caption{\textcolor{black}{Results on the multi-modal prostate DECATHLON dataset, with U-Net as backbone architecture. Best results are highlighted in bold.}}
\centering
\tiny
\begin{tabular}{ll|cccc|cc}
\toprule
                           &             & \multicolumn{2}{c}{\bf {T2}}                                 & \multicolumn{2}{c|}{\bf {ADC}}                            & \multicolumn{2}{c}{\bf {Fused}}                         \\ \midrule
                           & \bf {Method}              & \multicolumn{1}{c}{\bf {DSC}}      & \multicolumn{1}{c}{\bf {ASSD}} & \multicolumn{1}{c}{\bf {DSC}} & \multicolumn{1}{c|}{\bf {ASSD}} & \multicolumn{1}{c}{\bf {DSC}} & \multicolumn{1}{c}{\bf {ASSD}} \\ \midrule
\multirow{4}{*}{{CAM}}       & {Baseline}    & {32.89$\pm$8.34}               & {8.77$\pm$3.19 }           & {28.72$\pm$10.16}         & {15.73$\pm$5.15}            & {30.50$\pm$8.68}          & {13.48$\pm$3.93}           \\
                           & {SEAM(scale)\cite{wang2020self}} & {39.06$\pm$9.19}               & {9.12$\pm$3.28 }           & {52.67$\pm$11.92}         & {11.22$\pm$5.12 }           & {52.57$\pm$11.92}         & {11.21$\pm$5.17 }          \\
                           & {SEAM(all)\cite{wang2020self}}   & \bf {48.95$\pm$12.51}              & {9.32$\pm$5.93   }         & {62.65$\pm$11.37  }       & {5.97$\pm$4.64}             & {65.94$\pm$13.35}         & {6.30$\pm$4.16  }          \\
                           & {Proposed}    & {47.03$\pm$11.62} & \bf {6.20$\pm$3.61}            & \bf {70.51$\pm$13.41}         & \bf {2.95$\pm$1.81  }           & \bf {71.26$\pm$14.56}         & \bf {4.30$\pm$2.52  }          \\ \midrule
\multirow{4}{*}{{GradCAM++}} & {Baseline}    & {32.60$\pm$8.00 }              & {8.28$\pm$2.80}            & {28.71$\pm$10.25}         & {15.83$\pm$5.16}            & {30.76$\pm$8.75}          & {13.24$\pm$3.88}           \\
                           & {SEAM(scale)\cite{wang2020self}} & {39.07$\pm$9.18 }              & {9.11$\pm$3.27  }          & {52.69$\pm$11.91}         & {11.21$\pm$5.12  }          & {52.68$\pm$11.92 }        & {11.21$\pm$5.12  }         \\
                           & {SEAM(all)\cite{wang2020self}}   & \bf {48.94$\pm$12.50 }             & {9.32$\pm$5.92}            & {62.66$\pm$11.38}         & {5.96$\pm$4.64 }            & {65.75$\pm$13.22}         & {6.34$\pm$4.06 }           \\
                           & {Proposed}    & {47.85$\pm$11.14}              & \bf {6.47$\pm$3.39  }          & \bf {70.19$\pm$13.35}         & \bf {2.97$\pm$1.83  }           & \bf {71.10$\pm$14.62}         & \bf {4.23$\pm$2.58  }          \\ \midrule
{Upper Bound   }             &             & {86.82$\pm$2.88 }              & {1.06$\pm$0.29 }           & {78.80$\pm$6.97}          & {1.75$\pm$0.916}            &   {--}                      &      {--}                    \\ \bottomrule
\end{tabular}
\label{table:prostate}
\end{table*}

\textcolor{black}{We empirically demonstrate the generalization capabilities of our method by evaluating its performance on a different dataset, i.e., multi-modal segmentation from the prostate DECATHLON challenge. Table \ref{table:prostate} reports the results for both the baselines and the proposed approach. First, we can see that differences with respect to standard CAM and GradCAM++ methods are significant, with a margin of 15\% in MR-T2 and nearly 40\% in the ADC modality in terms of DSC. Regarding the values on the ASSD metric, we can also observe a substantial improvement, particularly in ADC. Second, compared to the prior state-of-the-art in \cite{wang2020self}, our method also brings an important gain in performance. Note that the original SEAM approach (i.e., \textit{SEAM(Scale)}) only integrated scale transformations as equivariant constraints, which underperformed the proposed approach by 8\% to 18\% in DSC and 3\% to 8\% in terms on ASSD. Last, compared to the enhanced version of SEAM \cite{wang2020self}, which incorporates multiple affine transformations, our multi-modal learning strategy yields considerable better results across all the metrics and scenarios, except in the MR-T2 modality and DSC, where both obtain similar performances.}

\begin{figure}[h!]
\begin{centering}
\includegraphics[width=1\linewidth]{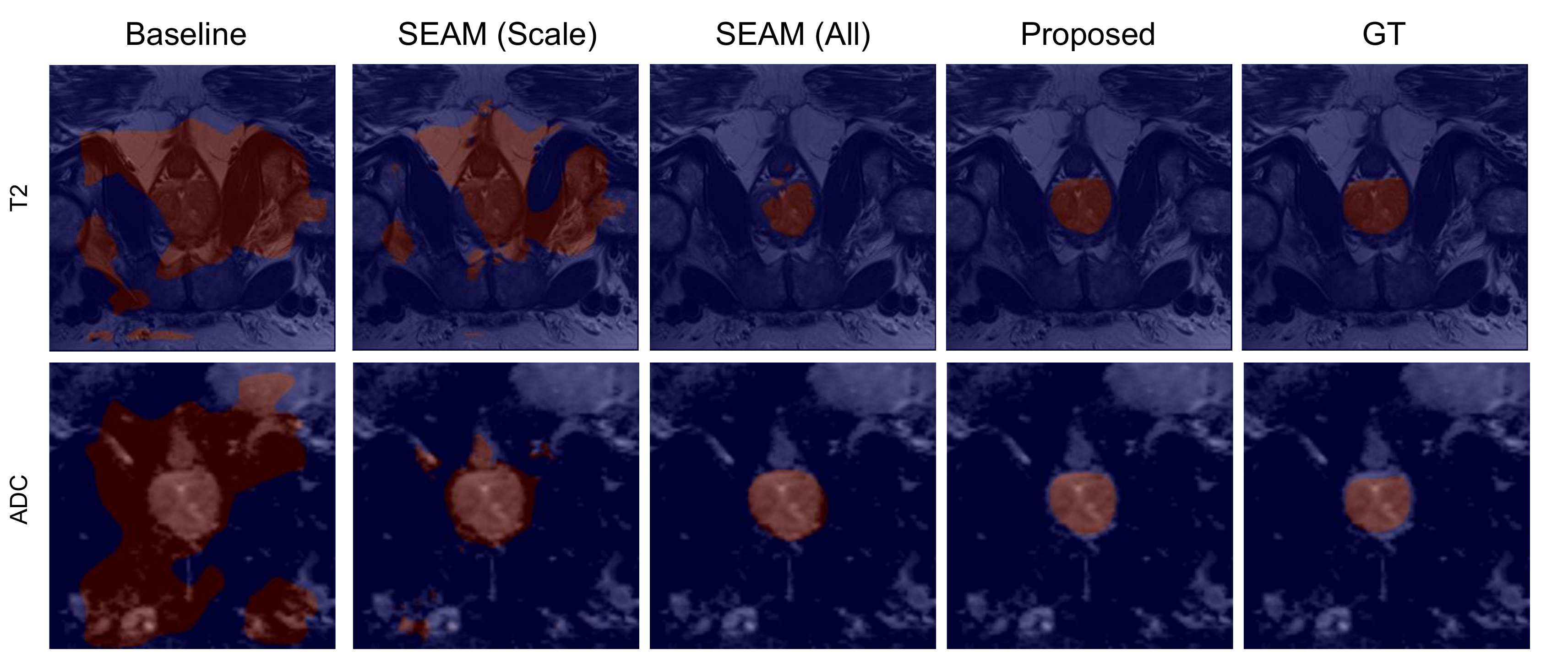}
\par\end{centering}
\caption{\textcolor{black}{Qualitative results on the prostate DECATHLON dataset on one 2D slide from a single validation scan (the binarized CAM is overlaid on the original image). Rows represent the image modality whereas the different models, as well as the corresponding ground truth are shown in the columns.}}
 
\label{fig:visual-results-prostate}
\end{figure}

These quantitative results are supported visually by the qualitative performance depicted in Figure \ref{fig:visual-results-prostate}. In these images, we can clearly see that both baseline and prior original work \cite{wang2020self} generate useless predictions. On the other hand, while integrating multiple affine transformations on SEAM results in better segmentations, these are arguably less accurate compared to the ground truth than the segmentations obtained by our approach. For example, the \textit{top row} shows that SEAM(All) produces significant undersegmentations, whereas the area identified in the \textit{bottom row} is larger than the ground truth. In contrast, the segmentations obtained by the proposed model resemble more closely the ground truth annotations, and are more consistent across modalities. These results suggest that our method can generalize efficiently to diverse multi-modal scenarios, while still outperforming existing approaches.

\subsection{{Impact of early concatenation}}

In this section, we demonstrate the benefits of the proposed model compared to using an early fusion strategy of multi-modal features through a single network with multi-channel inputs. More concretely, multiple modalities are simply concatenated at the input of the network, which contains $n$ channels, and where each channel corresponds to one modality. These results, which are reported in Table \ref{table:early_ours}, indicate that despite outperforming their singe-modality counterparts, baseline and prior state-of-the-art models do not leverage multi-modal data as efficiently as the proposed model. In particular, if cross-modality self-supervision is adopted, as in our approach, differences in DSC range from approximately 2\% to 7\%in the BraTS dataset, and nearly 30\% in the prostate DECATHLON data. We argue that the main limitation with this early-fusion setting is that, as the network generates the CAMs directly from the concatenation of $n$ modalities, we cannot implicitly leverage cross-modality constraints, which has shown to be helpful when learning common representations.

\begin{table*}[]
\caption{\textcolor{black}{Ablation study to assess the impact of early modality concatenation at the input on prior works. The last column represents the results obtained with our cross-modality strategy, where constraints are enforced at the output level. $\nabla$ represents the performance improvement compared to the baseline model, i.e., standard CAM and GradCAM++ from a network whose input is the concatenation of $n$ modalities. Best results highlighted in bold.}}
\centering
\tiny
\begin{tabular}{ll|cccccccc|cc}
\toprule
         &                 & \multicolumn{2}{c}{\bf {Baseline}}    & \multicolumn{2}{c}{\bf {SEAM(scale)}}                                           & \multicolumn{2}{c}{\bf {SEAM(all)}}                                             & \multicolumn{2}{c}{\bf {Proposed}}                     \\ \midrule
         & \bf {Combination}     & \bf {CAM}                                & \bf {Grad-CAM++}                         & \bf {CAM}                                 & \bf {Grad-CAM++}                          & \bf {CAM}         & \bf {Grad-CAM++}                          & \bf {CAM}                                 & \bf  {Grad-CAM++}      & {\bf $\nabla_{CAM}$} &  {\bf $\nabla_{Grad-CAM++}$}    \\ \midrule
\bf  {BraTS}    & {T1-T2}           &  {52.49$\pm$11.96}   &  {53.22$\pm$12.37} &  {54.35$\pm$11.86}   &  {56.53$\pm$12.36} &  {53.36$\pm$13.51}   &  {56.80$\pm$13.21} & \bf  {59.72$\pm$11.77}   & \bf  {59.46$\pm$11.90} &  {+7.23} &  {+6.24}\\
         &  {T1c-Flair}    &  {52.67$\pm$13.76}   &  {53.84$\pm$13.13} &  {50.41$\pm$14.02} &  {52.03$\pm$13.56}  &  {53.81$\pm$13.32}&  {55.25$\pm$12.74}  & \bf  {57.59$\pm$15.00} & \bf  {57.55$\pm$15.00} &  {+4.92} &  {+3.71}\\
         &  {T1-Flair}  &  {55.85$\pm$13.54}   &  {56.24$\pm$13.18}     &  {55.77$\pm$13.11}  &  {55.69$\pm$12.97} &  {56.38$\pm$12.66} &  {57.17$\pm$12.39} & \bf  {58.87$\pm$15.04} & \bf  {58.53$\pm$15.09} &  {+3.02} &  {+2.29}\\
         &  {T1c-T2}  &  {55.21$\pm$15.24} &  {56.23$\pm$15.13}                    &  {54.91$\pm$15.02}                     &  {54.42$\pm$15.33}                     &  {56.89$\pm$14.55}                     &  {56.26$\pm$15.53}                     & \bf  {59.16$\pm$11.33}                     & \bf  {58.93$\pm$13.41}  &  {+3.95} &  {+2.70}\\
         &  {T1-T2-T1c-Flair} &  {53.68$\pm$15.18}                    &  {53.73$\pm$14.89}                   &  {54.28$\pm$15.71}                     &  {54.32$\pm$15.40}                     &  {55.43$\pm$15.72}                     &  {55.56$\pm$15.67}                     & \bf  {56.81$\pm$13.84}                     & \bf  {57.05$\pm$14.06}  &  {+3.13} &  {+3.32}\\ \midrule
\bf  {Prostate} &  {T2-ADC}          & \multicolumn{1}{c}{ {40.77$\pm$8.17}} & \multicolumn{1}{c}{ {40.35$\pm$6.85}} & \multicolumn{1}{c}{ {46.10$\pm$12.01}} & \multicolumn{1}{c}{ {44.39$\pm$13.24}} & \multicolumn{1}{c}{ {66.74$\pm$14.58}} & \multicolumn{1}{c}{ {64.76$\pm$17.26}} & \multicolumn{1}{c}{\bf  {71.26$\pm$14.56}} & \bf  {71.10$\pm$14.62}  &  {+30.49} &  {+30.75}\\ \bottomrule
\end{tabular}
\label{table:early_ours}
\end{table*}

\subsection{ {Limitations}}

\textcolor{black}{Despite the empirical validation has demonstrated that the proposed approach can substantially improve the baselines performance, it presents several limitations. First, the whole learning objective can be only applied on multi-modal images, as the cross-modality equivariant constraint typically brings an important gain on performance. This does not prevent the proposed approach to be employed in a single-modality scenario. Nevertheless, one of the main contributions of this work is leveraging the information flow between image modalities by means of two learning objectives (i.e., cross-modality equivariant on CAMs and Kullback-Leibler divergence on the outputs). }

Second, the presented method is built upon the assumption that multi-modal images are co-aligned, which might not be the case in several scenarios. In particular, the proposed equivariant constraints on class activation maps leverage pixel-wise correspondences, which would not be applicable if images are misaligned. It is noteworthy to mention, however, that unpaired multi-modal segmentation has been indeed overlooked in the literature, even in the fully-supervised learning paradigm, with very few attempts to tackle this challenging setting \cite{multimodalKD}. Thus, we believe that tackling jointly weakly supervised learning on unpaired multi-modal data poses additional challenges that are worth to explore in the future. 

Furthermore, we have demonstrated that our learning strategy can be applied to diverse multi-modal scenarios, so that it can be generalized to multiple applications. Nevertheless, there exist some clinically relevant multi-modal data, such as PET-CT images, where the target might be easily visible in one modality but hardly distinguishable in the other. This may difficult the generation of class activation maps on the modality with low visible evidence of the target class, which could result in suboptimal joint CAMs. 

And last, we want to emphasize that we do not intend to solve the segmentation problem under the weakly supervised paradigm, but to pave the way towards closing the gap between weakly supervised and its fully supervised counterpart. We are aware that differences between both learning strategies are important. We want to emphasize that in this work we do not intend to solve the segmentation problem under the weakly supervised paradigm, but to pave the way towards closing the gap between weakly and fully supervised learning. If we look at our reported results, we can observe that the proposed approach brings a substantial performance gain compared to vanilla CAMs and existing approaches when they are trained under the exact same amount of supervision. The reported results align with most current literature under the learning with limited supervision paradigm. For example, recent works on weakly supervised or few-shot segmentation produce results that are far from full-supervision, with DSC values typically below 50$\%$ \cite{schumacher2020weakly,roy2020squeeze}. Thus, we believe that even though our model underperforms the fully supervised upperbound, it will be of broad interest to potentially close this gap, particularly in the multi-modal image scenario.

\section{Conclusion}
We proposed a novel learning strategy for multi-modal image segmentation under the weakly supervised learning paradigm. Whereas previous methods have overlooked the use of complimentary information across modalities, here we leverage their commonalities through the integration of explicit equivariant constraints. Our experiments show that the proposed method obtains better class activation maps than prior state-of-the-art approaches, \textcolor{black}{with extensive experiments on the impact of each component in our learning strategy, as well as results on two popular architectures. In addition, we have demonstrated the generabilization capabilities of the proposed approach by evaluating its performance in two well-known public datasets, demonstrating that our model consistently outperforms prior literature in these tasks.} Furthermore, the source of improvement is analyzed, proving that our approach generates better CAMs by means of reducing over-activated pixels while covering more complete regions. \textcolor{black}{Last, we have identified a set of potential limitations of our method, which can serve as basis for further improvements towards similar directions.} We believe this work might raise the interest on novel learning approaches to better model intra-modality information in segmentation neural networks, particularly in the medical domain, where multi-modal imaging is a common scenario.

\section*{Acknowledgments}
We would like to acknowledge Compute Canada for providing the computing resources employed in this work.

{
    % \clearpage
    \small
    \bibliographystyle{ieee_fullname}
    \bibliography{refs}
}

\end{document}